\documentclass{SCIS2025}

\usepackage[T1]{fontenc}
\usepackage{url} 
\usepackage{epigraph}
\usepackage{listings}
\usepackage{float}
\usepackage{xcolor}
\definecolor{changemark}{RGB}{0,0,0}
\usepackage{caption}
\usepackage{enumitem}
\usepackage{tcolorbox}
\usepackage{fontawesome5}
\usepackage{amssymb}       
\usepackage{pifont}        
\usepackage{colortbl}
\tcbuselibrary{listings,skins,breakable}
\usepackage{tikz}
\usetikzlibrary{arrows.meta,positioning,shapes,fit,backgrounds}
\IfFileExists{algorithm.sty}{\usepackage{algorithm}}{}
\usepackage{placeins}
\usepackage{adjustbox}
\usepackage{booktabs}
\usepackage{tabularx}
\IfFileExists{bbm.sty}{
  \usepackage{bbm}
}{
  \IfFileExists{dsfont.sty}{
    \usepackage{dsfont}
    \newcommand{\mathbbm}[1]{\mathds{#1}}
  }{
    \newcommand{\mathbbm}[1]{\mathbf{#1}}
  }
}

\lstdefinestyle{chatformat}{
    basicstyle=\ttfamily\footnotesize,
    breaklines=true,
    breakatwhitespace=false,
    backgroundcolor=\color{gray!5},
    frame=single,
    rulecolor=\color{gray},
    columns=fullflexible,
    keepspaces=true,
    aboveskip=1em,
    belowskip=1em,
    xleftmargin=0.5em,
    xrightmargin=0.5em,
}

\lstdefinelanguage{json}{
    morestring=[b]",
    showstringspaces=false,
    morecomment=[l]{//},
    breaklines=true,
    breakatwhitespace=true,
}

\lstdefinestyle{jsonstyle}{
    language=json,
    basicstyle=\ttfamily\footnotesize,
    columns=fullflexible,
    keepspaces=true,
    showstringspaces=false,
    breaklines=true,
    breakatwhitespace=true,
    aboveskip=0.5em,
    belowskip=0.5em,
}

\lstdefinestyle{pythonstyle}{
    language=Python,
    basicstyle=\ttfamily\footnotesize,
    numbers=left,
    numberstyle=\tiny,
    stepnumber=1,
    numbersep=8pt,
    columns=fullflexible,
    keepspaces=true,
    showstringspaces=false,
    breaklines=true,
    breakatwhitespace=true,
}

\definecolor{codebg}{rgb}{0.98,0.98,0.98}
\definecolor{codeframe}{rgb}{0.2,0.4,0.6}

\newtcblisting{prettypython}[1][]{
  enhanced,
  colback=codebg,
  colframe=codeframe,
  boxrule=1pt,       
  arc=2mm,           
  outer arc=1mm,
  fonttitle=\bfseries\color{white},
  coltitle=white,
  title filled=true,
  listing only,
  listing engine=listings,
  listing options={
    style=pythonstyle,
  },
  #1
}

\definecolor{systembg}{rgb}{0.95,0.95,0.95}
\definecolor{userbg}{rgb}{0.9,0.95,1.0}
\definecolor{assistantbg}{rgb}{0.95,1.0,0.95}
\definecolor{pythonbg}{rgb}{1.0,0.95,0.9}
\definecolor{promptborder}{rgb}{0.6,0.2,0.8}

\definecolor{myblue}{HTML}{9EDAE3}
\definecolor{mypink}{HTML}{E2B4BB}
\definecolor{myyellow}{HTML}{EED47E}
\definecolor{mygreen}{HTML}{A9C6A8}
\definecolor{mypurple}{HTML}{ADA8C6}

\newtcolorbox{systemmessage}{
  enhanced,
  colback=systembg,
  colframe=gray!50!black,
  arc=0.5mm,
  boxrule=0.5pt,
  width=0.9\linewidth,
  left=3mm,
  right=3mm,
  top=2mm,
  bottom=1mm,
  fontupper=\small,
  before upper={
  {\begingroup\bfseries\small\faCog\ \underline{System}\par\medskip\endgroup}
},
  frame code={
    \draw[gray!50!black, dashed, line width=0.5pt, rounded corners=1pt]
      (frame.south west) rectangle (frame.north east);
  }
}

\newtcolorbox{usermessage}{
  enhanced,
  colback=myblue,
  colframe=myblue!50!black,
  arc=0.5mm,
  boxrule=0.5pt,
  left=3mm,
  right=3mm,
  top=2mm,
  bottom=1mm,
  fonttitle=\bfseries\small,
  fontupper=\small,
  width=0.9\linewidth,
  coltitle=black,
  before upper={
  {\begingroup\bfseries\small\faUser\ \underline{User}\par\medskip\endgroup}
},
  frame code={
    \draw[myblue!50!black, dashed, line width=0.5pt, rounded corners=1pt]
      (frame.south west) rectangle (frame.north east);
  }
}

\newtcolorbox{assistantmessage}{
  enhanced,
  colback=myyellow,
  colframe=myyellow!50!black,
  arc=0.5mm,
  boxrule=0.5pt,
  left=3mm,
  right=3mm,
  top=2mm,
  bottom=1mm,
  fonttitle=\bfseries\small,
  fontupper=\small,
  width=0.9\linewidth,
  coltitle=black,
  before upper={
  {\begingroup\bfseries\small\faTools\ \underline{Tool Response}\par\medskip\endgroup}
},
  frame code={
    \draw[myyellow!50!black, dashed, line width=0.5pt, rounded corners=1pt]
      (frame.south west) rectangle (frame.north east);
  }
}

\newtcblisting{pythonresponse}{
  enhanced,
  colback=mypink!75,
  colframe=mypink!50!black,
  arc=0.5mm,
  boxrule=0.5pt,
  left=3mm,
  right=3mm,
  top=2mm,
  bottom=1mm,
  fonttitle=\bfseries\footnotesize,
  fontupper=\footnotesize,
  width=0.9\linewidth,
  coltitle=black,
  before upper={
    {\begingroup\bfseries\small\faRobot\ \underline{Assistant}\par\endgroup}

},
  frame code={
    \draw[myyellow!50!black, dashed, line width=0.5pt, rounded corners=1pt]
      (frame.south west) rectangle (frame.north east);
  },
  listing only,
  listing engine=listings,
  listing options={
    style=jsonstyle,
  }
}
\newtcblisting{completion}{
  enhanced,
  colback=white, 
  colframe=white, 
  boxrule=0pt, 
  arc=0mm, 
  left=0mm, 
  right=0mm, 
  top=0mm, 
  bottom=0mm, 
  title={\small\faRobot\space Assistant}, 
  fonttitle=\bfseries\small,
  fontupper=\small,
  before upper={\small},  
  width=0.9\linewidth,
  coltitle=black,
  attach boxed title to top left={yshift=0mm, xshift=0mm}, 
  boxed title style={
    colback=white,
    colframe=brown!70!black, 
    boxrule=0.5pt, 
    arc=1mm, 
    left=2mm,
    right=2mm
  },
  listing only,
  listing engine=listings,
  listing options={
    style=jsonstyle,
    basicstyle=\ttfamily\small,
  }
}

\usepackage[edges]{forest}
\definecolor{lightcoral}{rgb}{0.94, 0.5, 0.5}
\definecolor{lightgreen}{rgb}{0.56, 0.93, 0.56}
\definecolor{harvestgold}{rgb}{0.98, 0.85, 0.40}
\definecolor{brightlavender}{rgb}{0.75, 0.58, 0.89}
\definecolor{capri}{rgb}{0.0, 0.75, 1.0}
\definecolor{carminepink}{rgb}{0.92, 0.3, 0.26}
\definecolor{celadon}{rgb}{0.67, 0.88, 0.69}
\definecolor{darkpastelgreen}{rgb}{0.01, 0.75, 0.24}

\definecolor{hidden-draw}{RGB}{205, 44, 36}
\definecolor{hidden-blue}{RGB}{194,232,247}
\definecolor{hidden-orange}{RGB}{243,202,120}
\definecolor{hidden-yellow}{RGB}{242,244,193}
\definecolor{tree-level-1}{RGB}{245,20,85}
\definecolor{tree-level-2}{RGB}{246,86,118}
\definecolor{tree-level-3}{RGB}{248,177,193}
\definecolor{tree-leaf}{RGB}{176,230,198}

\definecolor{Self}{RGB}{255,0,128}
\definecolor{Ensemble}{RGB}{0,127,255}
\definecolor{Iterative}{RGB}{153,51,255}

\definecolor{exemplar1}{RGB}{136,98,148}
\definecolor{exemplar2}{RGB}{148,210,242}
\definecolor{knowledge1}{RGB}{249,219,152}
\definecolor{knowledge2}{RGB}{255,245,220}

\begin{document}
\ArticleType{Review}
\Year{2025}
\Month{}
\Vol{68}
\No{}
\DOI{}
\ArtNo{000000}
\ReceiveDate{}
\ReviseDate{}
\AcceptDate{}
\OnlineDate{}
\AuthorMark{}
\AuthorCitation{}

\title{Scaffolded Language Models with Language Supervision for Mixed-Autonomy: A Survey}{Scaffolded Language Models with Language Supervision for Mixed-Autonomy: A Survey}

\author[1]{Matthieu Lin}{}
\author[1]{Jenny Sheng}{}
\author[2]{Andrew Zhao}{}
\author[2]{\\Shenzhi Wang}{}
\author[2]{Yang Yue}{}
\author[3]{Victor Shea-Jay Huang}{}
\author[4]{Huan Liu}{}
\author[4]{Jun Liu}{}
\author[2]{\\Gao Huang}{}
\author[1]{Yong-Jin Liu}{liuyongjin@tsinghua.edu.cn}

\address[1]{Department of Computer Science and Technology, Tsinghua University, Beijing, China}
\address[2]{Department of Automation, Tsinghua University, Beijing, China}
\address[3]{Shanghai AI Lab, Shanghai, China}
\address[4]{Xi'an Jiaotong University, Xi'an, China}

\abstract{
This survey organizes the intricate literature on the design and optimization of emerging structures around post-trained LMs.
We refer to this overarching structure as scaffolded LMs and focus on \emph{a scaffold that integrates LMs in a multi-step process with tools}.
We view scaffolded LMs as semi-parametric models wherein we \emph{optimize prompts and tools, which we refer to as non-parametric variables}.
In particular, scaffolded LMs interpret instruction, use tools, and receive feedback all in language.
Recent works use an \emph{LM as an optimizer} to interpret language supervision and update non-parametric variables according to intricate objectives.
In this survey, we refer to this paradigm as \emph{training of scaffolded LMs with language supervision}.
A key feature of non-parametric training is the ability to \emph{learn from language}.
Parametric training excels in learning from demonstration (supervised learning), exploration (reinforcement learning), or observations (unsupervised learning), using well-defined loss functions.
Optimization in the space of language enables \emph{rich, interpretable, and expressive objectives}, while mitigating issues like catastrophic forgetting and supporting compatibility with closed-source models.
Furthermore, agents are increasingly deployed as co-workers in real-world applications—such as Copilot in Office tools or software development.
In these \emph{mixed-autonomy} settings, where control and decision-making are shared between humans and AI, users provide feedback by identifying errors or offering corrections.
{\color{changemark}Accordingly, we discuss agents that inhabit streams of experience \emph{by learning from this language-based feedback}.
}}

\keywords{Compound AI Systems, Language Agent, Mixed-Autonomy, Prompt Optimization, Experiential Learning Agents}

\maketitle

\section{Introduction}
\begin{figure}[ht]
    \centering
    \vspace{7pt}
    \includegraphics[width=1\linewidth]{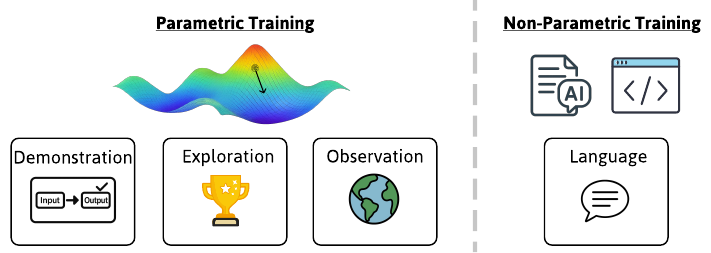}
    \caption[\textbf{Non-parametric training of scaffolded LMs focuses on learning from language.}]{\textbf{Non-parametric training of scaffolded LMs focuses on learning from language.} Parametric training has excelled in learning from demonstration (supervised learning), exploration (reinforcement learning), and observation (unsupervised learning) by using well-defined loss functions. Instead, non-parametric training enables using an LM to optimize in the language space, allowing for efficient and interpretable learning with fuzzy objectives and rich textual feedback. As scaffolded LMs interact with digital systems operated by humans, mixed autonomy becomes increasingly important \cite{mixedautonomy}. We anticipate that learning from language will serve as a crucial framework for scaffolded LMs that inhabit streams of experience from rich user feedback in mixed autonomy \cite{silver2025era}. Importantly, it circumvents catastrophic forgetting, supports compatibility with closed-source models, and is interpretable \cite{Hadsell2020}. We refer readers to Sec. \ref{sec:beyondiid} for further discussion.}
    \label{fig:1}
\end{figure}

Scaffolded language models embed language models (LMs) in a framework that extends their ability beyond traditional NLP tasks to perform open-ended digital automation tasks.
Notably, they mark the beginning of a new type of programs for digital automation \cite{openai_charter}.
It uses language as the interface between humans and digital systems.
This opens up the opportunity for mixed autonomy where humans and AI share control and decision-making responsibilities for digital tasks \cite{openhands, mixedautonomy}.

{\color{changemark}Following the terminology from OpenAI \cite{openai2024mlebench}, we define a scaffolded LM as a \emph{structure that integrates the LM into a multi-step process with tools}.
Specifically, the post-training interface of the LM enables the developer to specify the system behavior in natural language.
}%
The developer specifies available tools, and the LM executes them by generating the function calls and input arguments.
{\color{changemark}Tools \cite{openai_function_calling} correspond to external capabilities the model can invoke (see the glossary in Fig. \ref{fig:glossary}).
The scaffold refers to a surrounding structured code that manages control flow (multi-step process), maintains state (chat history), and executes tools based on responses of the LM.
Thus, the resulting scaffolded LM can be viewed as a semi-parametric model, where the LM is the parametric component and non-parametric components correspond to tools and prompts.
}%
In particular, non-parametric components enables an interpretable and efficient way to modify the behavior of the `program'.
For instance, we can add new or modify tools corresponding to APIs to interact with the digital world, developer-defined code to perform computations, or even other entities such as the human or another (scaffolded) LM \cite{wang2024what}.
As well as creating detailed prompts with intricate constraints.
The optimization of scaffolded LMs requires significant trial and error by expert developers.
This involves identifying the LM’s failure modes by running the scaffolded LM on examples of user queries and then addressing them one by one by refining the system prompt, input template, available tools, or the scaffold's code itself.

Accordingly, recent works \cite{textgrad, zhao2024expel} have proposed to use an LM as an optimizer that interprets intricate objectives, execution traces, and rich execution feedback to \emph{generate} updated non-parametric variables.
In practice, scaffolded LMs are naturally exposed to a wide variety of rich language feedback, such as user feedback or execution errors, that reveal about the failure modes.
Beyond learning from execution feedback, the LM optimizer can also extract insights or reusable tools from successful execution traces \cite{wang2025inducingprogrammaticskillsagentic, zhao2024expel}.

As shown in Fig. \ref{fig:1}, training of non-parametric variables focuses on learning from language.
Notably, parametric training excels in learning from demonstrations  (supervised learning) \cite{rlhf}, exploration (reinforcement learning) \cite{reasoning} , and observations (unsupervised learning)\cite{gpt3}, using well-defined loss functions \cite{lecun2015deep}.
In contrast, optimization in the language space \cite{llfbench} provides a learning system with intricate objectives expressed in natural language that is efficient and interpretable.
The training of non-parametric variables circumvents key limitations of gradient descent, such as catastrophic forgetting \cite{Hadsell2020}, while being compatible with closed-source models \cite{claude,chatgpt}.
In this survey, we refer to this framework as \emph{training of scaffolded language models with language supervision}.

This survey presents a taxonomy for understanding this paradigm shift from parametric training to non-parametric training.
In particular, it organizes the intricate literature on compound AI systems \cite{compound-ai-blog}, LM pipelines \cite{dspy}, prompt optimization \cite{ape, zhao2025compromisedprompts}, language agents \cite{openai2025agents, yaoshunyuthesis, rise-and-potential, landscape-emerging-agent}, AI workflow optimization \cite{trace, textgrad, adalflow}, and experiential learning agents \cite{zhao2024expel} under the concept of training of scaffolded language models with language supervision.
It focuses on the multi-step aspect that involves optimization beyond the prompt of a single LLM call \cite{ape} to multiple LM calls and tools.
Specifically, compound AI systems \cite{compound-ai-blog, lin2024llmbasedoptimizationcompoundai} focus on the systemic perspective (the interconnection of components) and overlook the multi-step aspect (\emph{agent vs workflow}).
Meanwhile, the literature in language agents \cite{rise-and-potential} neglects the efficiency of LM pipelines (also called workflows) \cite{dspy} that leverage the task structure to reduce the burden in decision-making and supervise the LM with clearer instructions.

{\color{changemark}As shown in Fig. \ref{fig:categorization_of_survey}, we organize this survey as follows:
\begin{enumerate}[label=\alph*)]
    \item Sec. \ref{sec:related_work} positions this survey in the broader literature of scaffolded LMs and recent breakthroughs in parametric training with RL.
    \item Sec. \ref{sec:scaffold} focuses on how the post-training interface of LMs interacts with the scaffold. We identify two types of scaffold—agents and workflows—and focus on important design aspects of each. Finally, we reflect on the trend of recent benchmarks and highlight the ability of scaffolded LMs for mixed-autonomy, a setting where human and AI share control and decision-making.
    \item Sec. \ref{sec:optimization} organizes the intricate literature that uses an LM as an optimizer for non-parametric training. We categorize these methods into whether they train LMs, agents, or workflows.
    \item Sec. \ref{sec:beyondiid} discusses a key capability missing from current scaffolded LMs: the ability to inhabit streams of experience \cite{silver2025era, yao2025second} and describes opportunities with non-parametric training.
\end{enumerate}
Additionally, to help readers navigate this survey, we provide a glossary in Fig. \ref{fig:glossary}.

\begin{figure}[ht]
    \centering
    \vspace{7pt}
    \scalebox{0.85}{\begin{forest}
        forked edges,
        for tree={
            grow=east,
            reversed=true,
            anchor=base west,
            parent anchor=east,
            child anchor=west,
            base=left,
            font=\small,
            rectangle,
            draw=hidden-draw,
            rounded corners,
            align=left,
            minimum width=4em,
            edge+={darkgray, line width=1pt},
            s sep=10pt,
            l sep=10pt,
            inner xsep=2pt,
            inner ysep=3pt,
            ver/.style={rotate=90, child anchor=north, parent anchor=south, anchor=center}
        },
        where level=1{text width=14em,font=\scriptsize,}{},
        where level=2{text width=14em,font=\scriptsize,}{},
        where level=3{text width=10em,font=\scriptsize,}{},
        [
            Training of Scaffolded Language Models With Language Supervision, ver, color=mypink!75!black, fill=mypink!100, text=black
            [
                Related Works \S \ref{sec:related_work}, color=mypurple!75!black, fill=mypurple!100, text=black
                [
                    Autonomous Agents \S \ref{subsec:autonomous_agents}, color=mypurple!75!black, fill=mypurple!60, text=black
                ]
                [
                    Parametric Training \S \ref{subsec:parametric_training}, color=mypurple!75!black, fill=mypurple!60, text=black
                ]
            ]
            [
                Scaffolded Language Models \S \ref{sec:scaffold}, color=myblue!75!black, fill=myblue!100, text=black
                [
                    Post-Trained \\ Language Models \S \ref{sec:languagemodel}, color=myblue!75!black, fill=myblue!60,  text=black
                ]
                [
                    Tools \S \ref{sec:tools}, color=myblue!75!black, fill=myblue!60, text=black
                ]
                [
                    Multi-step Process \S \ref{sec:multistep}, color=myblue!75!black, fill=myblue!60, text=black
                    [
                        Agent \S \ref{sec:agent}, color=myblue!75!black, fill=myblue!60, text=black
                    ]
                    [
                        Workflow \S \ref{sec:workflow}, color=myblue!75!black, fill=myblue!60, text=black
                    ]
                ]
                [
                    Benchmarks \S \ref{sec:benchmark}, color=myblue!75!black, fill=myblue!60, text=black
                ]
            ]
            [
                Training with Language Supervision \S \ref{sec:optimization}, color=myyellow!75!black, fill=myyellow!100, text=black
                [
                    Prompt Optimization \S \ref{sec:promptoptim}, color=myyellow!75!black, fill=myyellow!60, text=black
                ]
                [
                    Experiential Learning \S \ref{sec:expel}, color=myyellow!75!black, fill=myyellow!60, text=black
                    [
                        Insights \S \ref{sec:insight}, color=myyellow!75!black, fill=myyellow!60, text=black
                    ]
                    [
                        Tools \S \ref{sec:toolopt},color=myyellow!75!black, fill=myyellow!60, text=black
                    ]
                ]
                [
                    AutoDiff frameworks \S \ref{sec:autodiff}, color=myyellow!75!black, fill=myyellow!60, text=black
                    [
                        Graph Abstraction \S \ref{sec:graphabstraction}, color=myyellow!75!black, fill=myyellow!60, text=black
                    ]
                    [
                        Graph Execution and \\ Optimization \S \ref{sec:trace},color=myyellow!75!black, fill=myyellow!60, text=black
                    ]
                ]
            ]
            [
                Beyond Episodic Learning \S \ref{sec:beyondiid}, color=mygreen!75!black, fill=mygreen!100, text=black
                [
                    Scalability \S \ref{subsec:scalability}, color=mygreen!75!black, fill=mygreen!40, text=black
                ]
                [
                    Plasticity and Stability \S \ref{subsec:robustness}, color=mygreen!75!black, fill=mygreen!40, text=black
                ]
                [
                    Efficiency \S \ref{subsec:efficiency}, color=mygreen!75!black, fill=mygreen!40, text=black
                ]
                [
                    Interpretability \S \ref{subsec:interpretability}, color=mygreen!75!black, fill=mygreen!40, text=black
                ]
            ]
        ]
    \end{forest}
    }
    \caption{\textbf{Organization of this survey}. This survey introduces a unifying paradigm centered on learning from language using the non-parametric variables of a scaffolded language model. Despite the complexity and fragmentation of the literature across domains such as language agents, prompt optimization, experiential learning, and compound AI systems, we adopt a deliberately simple and clear organizational structure. This simplicity is not a limitation, but a reflection of our effort to distill a coherent framework from a rapidly evolving field. It enables us to outline key research opportunities in learning with non-parametric variables in mixed-autonomy settings.}
    \label{fig:categorization_of_survey}
\end{figure}

\begin{figure}[tbp]
    \centering
    \begin{adjustbox}{width=\textwidth}
    \begin{tcolorbox}[
      enhanced,
      colback=white,
      colframe=gray!40,
      arc=1mm,
      boxrule=0.6pt,
      left=2mm,right=2mm,top=2mm,bottom=2mm
    ]
    {\renewcommand{\arraystretch}{1.15}\small
    \rowcolors{2}{gray!6}{white}
    \arrayrulecolor{gray!55}\setlength{\arrayrulewidth}{0.5pt}
    \begin{tabularx}{\textwidth}{>{\bfseries}p{0.28\textwidth} X}
        \rowcolor{gray!15}
        Term & Definition \\
        \toprule
        \textcolor{changemark}{Scaffold (See Sec. \ref{sec:scaffold})} & \textcolor{changemark}{A surrounding structured code that manages control flow (multi-step process), maintains chat history, and executes tools based on responses of the LM.} \\
        \midrule
        \textcolor{changemark}{Chat format (See Sec. \ref{sec:languagemodel})} & \textcolor{changemark}{A chat format is a protocol that turns a structured conversation into a single sequence of strings that a post-trained LM both expects as input and emits as output.} \\
        \midrule
        \textcolor{changemark}{Tool (See Sec. \ref{sec:tools})} & \textcolor{changemark}{An external capability the model can invoke via the Tools API (e.g., developer-defined functions, search engine). A function is a specific tool defined by a JSON schema. Non-parametric training is used to implement or modify functions.} \\
        \midrule
        \textcolor{changemark}{Tool call (See Sec. \ref{sec:tools})} & \textcolor{changemark}{A specific type of response from the LM that includes a call to one of the available tools and a corresponding plain text for that tool (e.g., a query for a search engine); for a function call, it corresponds to a structured JSON.} \\
        \midrule
        \textcolor{changemark}{Mixed-autonomy (See Sec. \ref{sec:benchmark})} & \textcolor{changemark}{A setting where humans and AI work together, sharing control and decision-making responsibilities to achieve a goal. Typically, the human delegates control to the AI when appropriate, dividing the task according to their respective strengths. For instance, in vibe coding, the human guides intent and judges quality, while the AI proposes implementations and refactors.} \\
        \midrule
        \textcolor{changemark}{Prompt (See Sec. \ref{sec:optimization})} & \textcolor{changemark}{The prompt includes the system and user message that is provided to an LM at the start of the chat. The user message specifies the task and data for the episode. The system message is shared across tasks, it sets the system behavior by specifying available tools, role of the scaffolded LM, high-level instructions, constraints, and any other information required to perform the task. Non-parametric training is used to improve the system prompt, e.g., by adding high-level insights or improving tool description.} \\
        \midrule
        \textcolor{changemark}{Non-parametric variables (See Sec. \ref{sec:optimization})} & \textcolor{changemark}{Prompts and tools of a scaffolded LM. Agents have a single prompt while workflows use a different prompt at each step. Non-parametric variables are expressed either in language or code.} \\
        \midrule
        \textcolor{changemark}{LM optimizer (See Sec. \ref{sec:optimization})} & \textcolor{changemark}{An LM optimizer prompts an LM with an objective function expressed in natural language to generate updated non-parametric variables.} \\
        \midrule
        \textcolor{changemark}{Language supervision (See Sec. \ref{sec:optimization})} & \textcolor{changemark}{It refers to the objective function and execution feedback, both expressed in natural language, that are provided to the LM optimizer.} \\
        \midrule
        \textcolor{changemark}{Episode (See Sec. \ref{sec:beyondiid})} & \textcolor{changemark}{Each episode begins with a new chat history when the human delegates control to the scaffolded LM via an instruction and concludes once the scaffolded LM successfully completes the task or the human resumes control. Importantly, during an episode, the human can offer corrections or provide guidance. Its textual form corresponds to an execution trace and feedback, which is used to prompt the LM optimizer.} \\
        \bottomrule
    \end{tabularx}}
    \end{tcolorbox}
    \end{adjustbox}
    \caption[\textbf{Glossary of core terms.}]{\textcolor{changemark}{\textbf{Glossary of core terms.}}}
    \label{fig:glossary}
\end{figure}

\section{Related Works}
\label{sec:related_work}

This section clarifies the position of this survey in the broader literature of autonomous agents \cite{llm-autonomous-agents-survey, rise-and-potential,feedback-mechanism-llm-agents} and parametric training \cite{deepseekai2025deepseekr1incentivizingreasoningcapability}.

\subsection{Autonomous Agents}
\label{subsec:autonomous_agents}

In this survey, we deliberately adopt the broader \emph{scaffold} abstraction that encompasses both \emph{agents} and pre-structured \emph{workflows}.
This is unlike existing surveys that focus on \emph{agents} as the primary abstraction \cite{llm-autonomous-agents-survey, rise-and-potential,feedback-mechanism-llm-agents}.
In particular, they focus on systematizing components such as perception, planning, action loops, tool use, memory, and environment interaction \cite{llm-autonomous-agents-survey, rise-and-potential}, or feedback mechanisms (e.g., human preferences) to improve agent behavior \cite{feedback-mechanism-llm-agents}.
We separate \emph{learning paradigms} into (i) \emph{parametric} training that updates LM weights (e.g., supervised learning, RL) and (ii) \emph{non-parametric} training that updates non-parametric variables including the prompt and tools based on language-feedback.
We argue that this separation is crucial for mixed-autonomy settings: non-parametric updates are interpretable and easily supervised by humans at high frequency, and they synergize with parametric post-training to deliver data-efficient learning.
Using this taxonomy, we unify prompt optimization, experiential learning, and AutoDiff-style frameworks under a single lens.

\subsection{Parametric Training}
\label{subsec:parametric_training}

%

The discussion of parametric training \cite{tie2025posttraining, lai-etal-2025-survey, du2025surveyoptimizationlargelanguage} is important to understand the relevance of non-parametric training in the context of recent breakthrough in post-training of LMs with RL \cite{deepseekai2025deepseekr1incentivizingreasoningcapability,lai-etal-2025-survey,tie2025posttraining}.
We start with an example in prompt optimization where recent methods that use parametric training to learn to reason offer much better performance \cite{deepseekai2025deepseekr1incentivizingreasoningcapability}.
To understand this, we compare RL with verifiable reward \cite{deepseekai2025deepseekr1incentivizingreasoningcapability} with previous approaches.
Provided with a reward that measures the performance of a model on a downstream task by verifying that the model's answer match the ground-truth response \cite{cobbe2021gsm8k}:
\begin{equation}
    r \triangleq\mathbb{E}_{\mathbf{x}\sim \mathcal{D}}\Big[\mathbb{E}_{\mathbf{y}\sim p(\mathbf{x})}[
\mathbbm{1}\{\mathbf{y} = \mathbf{y}^{\star}(\mathbf{x})\}
]\Big],
\label{eq:downstream}
\end{equation}
previously the following approaches have been used.
One approach is to use an LM optimizer to generate a prompt and evaluate it against the reward.
These methods use search methods such as MCTS \cite{promptagent} or beam search \cite{protegi}.
Another approach train the LM optimizer with RL to generate better prompts \cite{zhao2024diverctdiversityenhancedredteaming, deng2022rlpromptoptimizingdiscretetext, prewrite, stableprompt}.
Similarly, another line of work \cite{tempera} proposes test-time editing of the instruction, resulting in a query-dependent LM optimizer.
Tempera \cite{tempera} uses a restricted action space (e.g., modify the verbalizer) to reduce variance in RL training.
They define the reward as the relative performance before and after the edit on the downstream task.
Recently, with stronger base models, LMs can directly optimize Eq. \ref{eq:downstream} with RL \cite{deepseekai2025deepseekr1incentivizingreasoningcapability}.
Importantly, doing so results in strong generalization, improvements beyond the domain at hand \cite{zhao2025absolutezeroreinforcedselfplay, wang2025reinforcementlearningreasoninglarge, wei2025swerl} and opens up a new axis for scaling, test-time compute.
Similar examples can be found in scaffolded LMs trained with RL.
SWE-RL \cite{wei2025swerl} compares the generated patch with the GT patch on SWE-Bench.
They do not use test cases as the reward signal, which can lead to reward hacking \cite{liu2023is}.
Specifically, SWE-RL is based on the Agentless \cite{agentless} and applies RL on the patch generation step.
They find that, unlike SFT, training with RL on SWE-Bench \cite{jimenez2024swebench} leads to improvement across domains, including code generation with library use \cite{zhuo2025bigcodebench}, code reasoning \cite{codereasoning}, math \cite{mathhendrycks}, and language understanding \cite{hendrycks2021measuring}.
Thus, for tasks that are easily verifiable, parametric training works very well  \cite{wei2025verifiersrule}.
However, as we discuss in Sec. \ref{sec:beyondiid}, non-parametric training offers interesting avenues of research in circumventing catastrophic forgetting and the plasticity-stability dilemma, allowing continuous learning across episodes \cite{silver2025era}.
}
\section{Scaffolded Language Model}
\label{sec:scaffold}

A scaffolded LM refers to an LM embedded in a multi-step process with tools.
{\color{changemark}Such a structure allows for extending an LM's utility to interact with digital systems in similar ways as humans do \cite{openhands}.
Scaffolded LMs involve tools and thus differs from meta-generation techniques \cite{welleck2024from} that employ multiple calls to an LM \cite{adas, zhang2025aflow, archon}.
%
Meta-generation mimics search algorithms \cite{silver2016mastering} for reasoning tasks \cite{cobbe2021gsm8k} where each step consists of a call to an LM.
Importantly, in scaffolded LMs, each call to an LM expects a tool response.
}

To provide readers with a practical understanding of how this structure allows LMs to perform sophisticated cognitive tasks and improve from non-parametric training, we organize this section as follows.
We introduce the post-training interface of language models that allows interaction with tools and defines their behavior in natural language.
%
{\color{changemark}Next, we discuss multiple aspects of tools including compositionality, reliability, safety, and built-in tools.
}%
Furthermore, we show how the multi-step process can be controlled to implement priors into how the task is solved.
Finally, we discuss benchmarks to understand the shift induced by scaffolded LMs.
{\color{changemark}Specifically, these benchmarks highlight the omnipresence of scaffolded LMs in mixed-autonomy settings, where the design of the scaffold defines the interaction dynamics between human and AI agents for sharing control and decision-making.
}
\subsection{Post-trained Language Models}
\label{sec:languagemodel}

\begin{figure}[ht]
  \centering
    \adjustbox{width=0.75\linewidth,keepaspectratio}{%
        \begin{tcolorbox}[
          enhanced,
          colback=white,
          colframe=black!50,
          arc=1.5mm,
          boxrule=1pt,
          fonttitle=\bfseries\small,
          coltitle=white,
          colbacktitle=promptborder,
          attach boxed title to top center={yshift=-3mm},
          watermark text={\raisebox{8cm}{\scalebox{5}{\rotatebox{45}{\textcolor{promptborder!10}{MODEL INPUT}}}}},
          watermark color=promptborder!10,
          width=0.9\linewidth,
        ]
        \centering
        \begin{systemmessage}
          You have access to the following functions. Respond in JSON format.

          \begin{tcolorbox}[
            listing only,
            listing engine=listings,
            colback=gray!5,
            colframe=gray!50!black,
            boxrule=0.3mm,
            arc=1mm,
            listing options={
              language=json,
              basicstyle=\ttfamily\footnotesize,
              breaklines=true
            }
          ]
\begin{lstlisting}[style=jsonstyle]
{
  "name": "search_web",
  "description": "Search the web for current information",
  "parameters": {
    "type": "object",
    "properties": {
      "query": {
        "type": "string",
        "description": "The search query"
      }
    },
    "required": ["query"]
  }
}
\end{lstlisting}
          \end{tcolorbox}

          You are an assistant that provides the user with information from the web.
        \end{systemmessage}

        \begin{usermessage}
          Find the latest trip of IShowSpeed
        \end{usermessage}

        \begin{pythonresponse}
{"name": "search_web", "parameters": {"query": "IShowSpeed latest trip"}}
        \end{pythonresponse}

        \begin{assistantmessage}
          IShowSpeed's most recent trip was a high-profile tour across China and Mongolia in March and April 2025.
        \end{assistantmessage}

        \end{tcolorbox}
    }
    \caption[Example of chat history in LLaMA-3.]{\textcolor{changemark}{Example of chat history in LLaMA-3.} Adapted from \cite{grattafiori2024llama3herdmodels}. A chat history starts with a prompt comprising a system and user message in LLaMA-3. The system message enumerates available tools (as JSON schemas) and sets high-level behavior. The user message specifies the task and the data. During the multi-step process, the chat history is updated with assistant messages (LM responses), user feedback, and tool responses.}
    \label{fig:chat}
\end{figure}

{\color{changemark}From a probabilistic perspective, given an input (corresponding to the chat history) $\mathbf{x} = (x_1, \dots, x_T)$ consisting of a sequence of tokens $x_t \in \mathcal{V}$, a post-trained LM generates a response $\mathbf{y} = (y_1, \dots, y_N)$ conditioned on $\mathbf{x}$ in an autoregressive manner:
}$$
p_{\theta}(\mathbf{y} \mid \mathbf{x}) = \prod_{t=1}^N p_{\theta}(y_t \mid \mathbf{x}, y_{<t}).
$$
{\color{changemark}During post-training, the LM is trained to follow a \emph{chat format}. Thus, the response follows the chat format and corresponds to a tool call or a response to the user.
As shown in Fig. \ref{fig:chat}, the multi-step process in a scaffolded LM produces a chat history where messages come from different roles.
Specifically, a chat format is a protocol that specifies how to convert a chat history into a single sequence of tokens $\mathbf{x}$ that the post-trained LM both expects as input and emits as output.
For instance, the ``tool'' role is used for tool responses.
In other words, we can understand a chat format as a protocol consisting of
\begin{itemize}[label=\arabic*.]
    \item a finite set of roles, e.g., system, user, assistant, tool.
    \item a set of special tokens that act as markers, such as role headers, channel indicators, message delimiters, tool-call fences, and termination markers \cite{kundel2025harmony}.
\end{itemize}
Importantly, the system message and user message specify how the model should behave as well as when and how to use tools.
A chat starts with a system message and a user message, and we refer to them as the \emph{prompt}.
Non-parametric training steers the behavior of scaffolded LMs by optimizing the prompt and tools.

By default, it is recommended to adhere to the provider's native chat format with explicit role headers (system, user, assistant, tool), placing tool outputs under the dedicated tool role and assistant generations under the assistant role.
This aligns with how post-trained LMs are supervised and preserves the intended privilege ordering \cite{wallace2024instructionhierarchytrainingllms}.
During post-training, the LM is explicitly trained to treat each message source (system → user → tool output) as having an explicit privilege order \cite{wallace2024instructionhierarchytrainingllms}.
Thus, if there's a conflict between messages from different sources, the LM is trained to prioritize them based on this privilege order.
This avoids prompt-injection \cite{openai_codex_cloud_internet_access}, jailbreaks \cite{chao2023jailbreaking}, and system-prompt extraction \cite{breunig2025claude}.
This process is different from knowledge conflict \cite{li2025tamingknowledgeconflictslanguage, knowledge-conflict-survey} that focuses on factuality.
It arises when there's a conflict between parametric knowledge (what the model “knows”), contextual knowledge (retrieved content), and instruction knowledge (user/system assumptions or edits).
%
%
Accordingly, Claude's system prompt includes explicit instructions to avoid invoking external tools on topics the model already ``knows,'' which could be a strategy to avoid knowledge conflicts to some extent \cite{breunig2025claude, asgeirtj2025claude}.
}

\subsection{Tools}
\label{sec:tools}

\begin{figure}[ht]
    \centering
    \vspace{7pt}
    \includegraphics[width=0.85\linewidth]{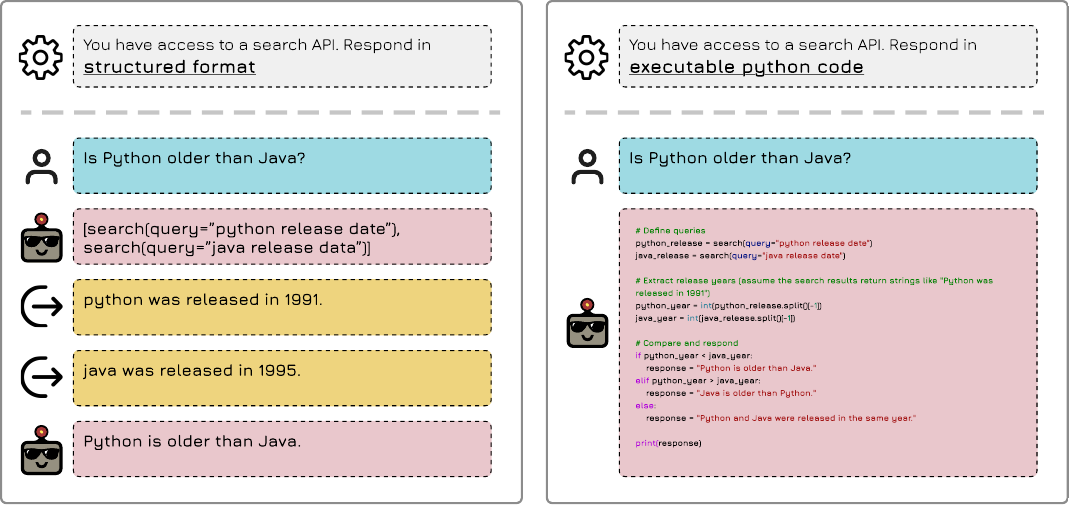}
    \caption[\textbf{Tool compositionality.}]{\textcolor{changemark}{\textbf{Tool compositionality.} Left: tools are composed sequentially. As shown in the figure, some LMs support generating independent tool calls in parallel \cite{openai_function_calling}. Right: executable Python code as a unified way to call tools \cite{codeact}. Code inherently supports both control flow (e.g., conditionals and loops) and data flow, allowing intermediate results to be stored and reused via variables when coordinating multiple tool calls}}
    \label{fig:tool}
\end{figure}

{\color{changemark}A tool can be virtually anything, including the interface to a digital system, a scratch pad for thinking \cite{anthropic2025thinktool}, a request to the user \cite{langchain_human_tool}, another LM \cite{openai_agents_sdk}, or developer-defined code.
We define a tool as an external capability that the LM can invoke via the tool API \cite{openai_function_calling, wang2024what}.
Moreover, we define a function as a tool defined by a JSON schema.
Similarly, we define tool calling as a specific type of response from the LM that includes a call to one of the available tools and a corresponding plain text for that tool (e.g., a query to a search engine).
We define function calling as a specific type of tool call that corresponds to a structured JSON.
It is important to note that the order of arguments in a function matters because post-trained LMs are autoregressive models \cite{mccoy2023embersautoregressionunderstandinglarge, grosse2023studyinglargelanguagemodel}.

During post-training, LMs may be trained to use built-in tools, which we call \emph{tool-integrated reasoning} \cite{openai2025gptoss120bgptoss20bmodel}. Built-in tools are executed remotely on the provider side within temporary secure sandboxes, rather than on the user's local machine. Thus, built-in tools are only used for seeking external information to improve their reasoning process rather than advancing on the task itself. Built-in tools \cite{openai2025agents, anthropic2025thinktool} typically include:
}\begin{itemize}
    \item a web search API \cite{nakano2022webgptbrowserassistedquestionansweringhuman} allows the LM to tackle tasks with public data. The LM typically generates a response by citing the retrieved content.
    \item a vector database \cite{rag, basic, ralm, context-retrieval, hover} for storing data as embeddings that can be retrieved through top-$k$ retrieval for a query. It is often used for accessing private data or as a memory module \cite{generative-agents}.
    \item a computer use tool \cite{openai2025operator} that interacts with a headful browser. Unlike the web search API, it allows the LM to use interactive websites.
\end{itemize}
{\color{changemark}

The multi-step process in scaffolded LMs naturally allows sequential tool composition, where tools are called sequentially in a coordinated manner to solve a task.
In that sense, scaffolded LMs start by producing a plan in plain text to improve tool composition \cite{openai2025gptoss120bgptoss20bmodel}.
CodeAct \cite{codeact} proposes to use executable Python code as a unified way to call tools.
As shown in Fig. \ref{fig:tool}, code inherently supports both control flow (e.g., conditionals and loops) and data flow, allowing intermediate results to be stored and reused via variables when coordinating multiple tool calls.
They \cite{codeact} find that it achieves comparable or better performance on sequential tool composition.
However, this approach is vulnerable to executing unsafe code and requires sandboxing.
Even with sandboxes, strict capability scoping and comprehensive controls are needed: network and file-system egress restrictions, timeouts, resource quotas, import allow-lists, AST/static analysis, and immutable audit trails.
Building and operating that infrastructure is non-trivial.
Moreover, code-injection attacks become more dangerous because untrusted tool outputs (e.g., web pages or file contents) can be stitched into code and executed.
Additionally, agents with access to the internet are exposed to a wide variety of attacks where tools return instructions, when followed by the LM, leads to leaking personal data \cite{openai_codex_cloud_internet_access}.

Besides, LMs may also occasionally generate invalid tool calls (e.g., syntax errors), wrong tools, or irrelevant tools.
For instance, irrelevant tool calls may be a result of reward hacking during post-training \cite{ma2025pouproofuse}.
Similarly, the LM may face unexpected tool execution failures (e.g., rate-limit).
To account for these, developers must implement error-handling mechanisms to catch exceptions during tool execution and provide feedback to the LM for recovery.
Additionally, the way tools are defined and described inside the system message affects reliability.
Clear, specific descriptions and intuitive names help the LM choose the right tool.
Additionally, clear description provides explicit guidance for the LM to learn when to use this tool as opposed to learning it through RL which is prone to reward hacking \cite{ma2025pouproofuse}.
Thus, the optimization of prompts and tools is important.
}

\subsection{Multi-Step Process}
\label{sec:multistep}
\begin{figure}[ht]
    \centering
    \includegraphics[width=0.85\linewidth]{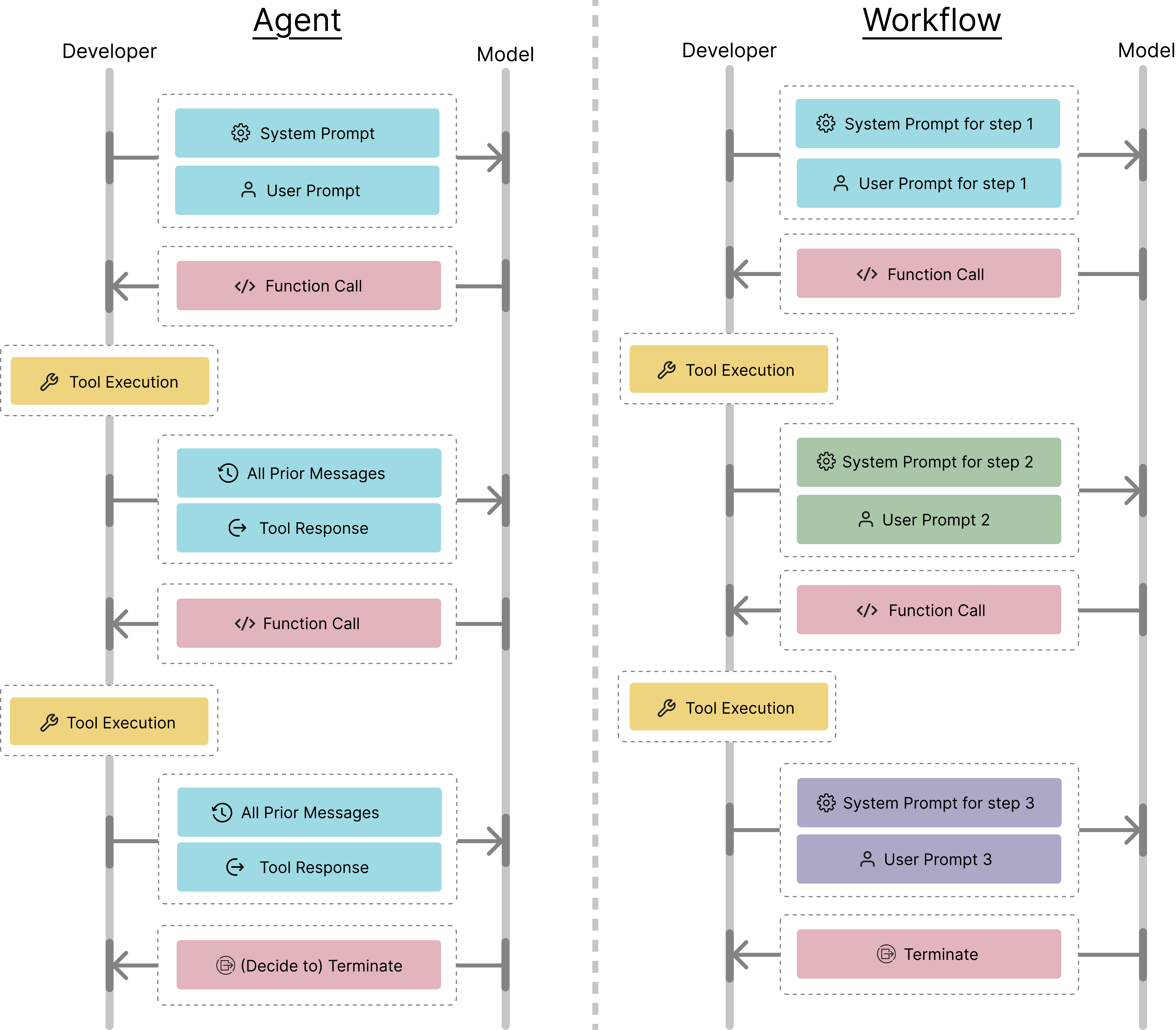}
    \caption[\textbf{Multi-step process of a scaffolded LM}.]{\textcolor{changemark}{\textbf{Multi-step process of a scaffolded LM}. The distinctive trait between agent and workflow is that for each workflow step, a different prompt is written to pre-define the sub-task at that step. This makes workflows much more efficient and predictable than agents. However, workflows are engineered for specific scenarios. In this figure, we assume the case where the tool execution happens on the developer side. For built-in tools, tool execution happens on the provider's side \cite{openai2025agents}.}}
    \label{fig:agent-workflow}
\end{figure}

{\color{changemark}A scaffolded LM involves a multi-step process with tools.
Each step consists of an interaction between the assistant and tool.
}%
As shown in Fig. \ref{fig:agent-workflow}, guiding the model at each step with tailored prompts reduces its decision-making burden.
{\color{changemark}Thus, in practice, a workflow involves pre-defining the number of steps and crafting appropriate prompt for each step, while an agent focuses on engineering a single prompt and the set of tools.
}

These approaches mainly differ in how much autonomy the LM has, and we can also refer to this autonomy as the degree of agenticness \cite{langchain2025agentframeworks}.
An agent works well in unstructured scenarios, i.e., in cases where it is hard to anticipate all the scenarios that the LM might face.
In contrast, a workflow approach works well when we know the solution path to a problem and can thus directly define each step required to solve it.
A good example is solving GitHub issues in SWE-Bench \cite{jimenez2024swebench}.
A workflow such as Agentless \cite{agentless} solves GitHub issues with the following steps.
Roughly, it first calls the LLM with the GitHub repository and issue description to provide a list of files that potentially contain the bug.
Those files are then used to prompt the LLM to choose relevant content inside those files.
Then, it asks the LLM to generate multiple patches.
In comparison, an agent such as Openhands \cite{openhands} provides the LM with a terminal and a code interpreter to interact with the repository.

\subsubsection{Agents}
\label{sec:agent}

Agents solve tasks by deciding at each step which tool to use and how to use it.
They also determine when a task is solved based on their own interpretation of results and goals.
{\color{changemark}The chat of an agent starts with a prompt and each step updates the chat history.
The agent continues until it decides to stop or it reaches the maximum number of steps, and returns a response to the user that summarize what it did.
Besides providing the right tool to our agent, an important design of the agent is how it composes tools.
In particular, code execution \cite{codeact} offers a unified way to compose tools with control flow and data flow, and typically results in a shorter horizon.
Another approach prompts the agent to provide a plan beforehand \cite{openai2025gptoss120bgptoss20bmodel}.
}

\subsubsection{Workflows}
\label{sec:workflow}

Workflows \cite{agentless, lu2024aiscientist, aiscientist_v2} shine in applications where a solution path can be pre-determined.
{\color{changemark}In contrast to agents, they are rendered more predictable and reliable by limiting the LM's autonomy.
At each step, the input $\mathbf{x}$ consists of a different prompt that specify the sub-task at that step.
}%
Thus, workflows consist of a pre-determined number of steps.
It is more cost-efficient as it avoids any unnecessary exploration steps \cite{agentless}.
\textbf{Unlike agents that scale with the number of steps} \cite{pan2024trainingsoftwareengineeringagents}, \textbf{workflows scale by generating multiple responses in parallel} \cite{snell2024scalingllmtesttimecompute, welleck2024from, chen2024llmcallsneedscaling}.
Specifically, by fixing the solution path, the output format at each step is predictable.
Thus, workflows use meta-generation techniques \cite{welleck2024from} such as best of $n$ sampling or majority voting \cite{self-consistency} to scale test-time compute.
For instance, in \cite{aiscientist_v2}, the AI scientist-v2 workflow can reach workshop-level papers by implementing tree-based methods on top of its previous version \cite{lu2024aiscientist}.
In particular, by generating multiple solutions at each step, it effectively generates multiple solution paths, and then uses heuristics (e.g., LLM as judge \cite{tot}) to select the best-performing path.
{\color{changemark}%
The cost of parallelization in workflows mainly depends on the aggregation method.
For instance LLM as judge adds one extra LM call, while pairwise judging adds quadratic cost.
Meanwhile, aggregation can be done without an LM judge, e.g., unit tests \cite{agentless} or majority voting \cite{self-consistency}.

}

\subsection{Benchmarks}
\label{sec:benchmark}

{\color{changemark}We highlight a growing trend in recent benchmarks: from measuring pure intelligence (e.g., math olympiads) to measuring productivity gains (e.g., software engineering) \cite{openai_charter}.
}%
The goal is no longer to assess whether a model solves a task perfectly in isolation, but how reliably and usefully it supports human workflows.
{\color{changemark}Importantly, these benchmarks are used to compare LMs on products \cite{windsurf_editor, openhands, cursor2023, microsoft2025copilot} that operate in mixed-autonomy.
This is a setting where humans and AI work together, sharing control and decision-making responsibilities to achieve a goal.
For instance, in vibe coding, the human guides intent and judges quality, while the AI proposes implementations and refactors.

To accurately gauge mixed-autonomy capability, these benchmarks align agent performance with human effort—measuring monetary return \cite{miserendino2025swelancerfrontierllmsearn}, simulating human collaborators \cite{tau-bench, zhou2025sweetrltrainingmultiturnllm}, exercising real software and APIs \cite{jimenez2024swebench, wei2025browsecompsimplechallengingbenchmark, starace2025paperbenchevaluatingaisability}, and using time as a measure of task difficulty \cite{rein2025hcasthumancalibratedautonomysoftware} because \emph{time is money}.

SWE-bench \cite{jimenez2024swebench} and SWE-Lancer \cite{miserendino2025swelancerfrontierllmsearn} ground evaluation in software engineering tasks from GitHub or Upwork—tasks that are compensated in real-world settings and evaluated via functional test cases.
Other benchmarks go beyond measuring raw accuracy.
BrowseComp \cite{wei2025browsecompsimplechallengingbenchmark} evaluates accuracy and calibration error.
In particular, the LM is asked to output a confidence (0–100) with its answer; BrowseComp then evaluates how well that confidence aligns with reality.
HCAST \cite{rein2025hcasthumancalibratedautonomysoftware} evaluates success rate as a function of human-calibrated task length (e.g., \textless 15min, 15min-15h, 1-4h, 4h+).
Additionally, they compute a ``50\% time-horizon'': the human task duration at which the agent has a 50\% chance of success, fitted from per-task success vs. human time.
RE-Bench \cite{wijk2024rebenchevaluatingfrontierai} finds that currently humans scale better with increased time budgets to solve the task.
For example, $\tau$-Bench \cite{tau-bench} and Sweet-RL \cite{zhou2025sweetrltrainingmultiturnllm} simulate human feedback using LMs to better reflect the back-and-forth nature of co-working.
For tasks that are hard to verify, PaperBench \cite{starace2025paperbenchevaluatingaisability} proposes a rubric organized as a hierarchical tree of requirements.
The leaf nodes are precise, binary checks (pass/fail).
Internal nodes aggregate their children by a manual, importance-based weighting, and the root score is the paper's Replication Score.
Tab. \ref{tab:benchmarks} summarizes these benchmarks.

\begin{table*}[t]
    \centering
    \begin{adjustbox}{max width=\textwidth}
    {\small
    \rowcolors{2}{gray!6}{white}
    \arrayrulecolor{gray!55}\setlength{\arrayrulewidth}{0.5pt}
    \begin{tabularx}{1.02\textwidth}{>{\raggedright\arraybackslash}p{0.30\textwidth} >{\raggedright\arraybackslash}p{0.28\textwidth} >{\raggedright\arraybackslash}X}
        \rowcolor{gray!15}
        \textbf{Benchmark} & \textbf{Scenario} & \textbf{Evaluation metrics} \\
        \toprule
        SWE-Bench \cite{jimenez2024swebench} & Software engineering & Unit test. \\
        SWE-Lancer \cite{miserendino2025swelancerfrontierllmsearn} & Software engineering & pass@1 / pass@k, dollars earned, unit test, SWE Manager by matching the original manager's proposal. \\
        HCAST \cite{rein2025hcasthumancalibratedautonomysoftware} & Machine learning, software engineering, cybersecurity & Binarized success rate by human-time buckets. \\
        RE-Bench \cite{wijk2024rebenchevaluatingfrontierai} & AI Research & Expert-graded task success; fraction of human pace; time-to-solution. \\
        PaperBench \cite{starace2025paperbenchevaluatingaisability} & AI research & Rubric organized as a hierarchical tree of requirements. \\
        $\tau$-Bench \cite{tau-bench} & Tool-agent-user interaction & pass@k metric as a measure of reliability of an agent. \\
        SWEET-RL \cite{zhou2025sweetrltrainingmultiturnllm} & Multi-turn collaborative reasoning & Task success/win rate under simulated feedback. \\
        BrowseComp \cite{wei2025browsecompsimplechallengingbenchmark} & Deep research & Accuracy, calibration error. \\

        \bottomrule
    \end{tabularx}}
    \end{adjustbox}
    \caption{\textbf{Benchmarks commonly used for scaffolded LMs}. }
    \label{tab:benchmarks}
\end{table*}

}
\section{Training with Language Supervision}
\label{sec:optimization}

\begin{figure}[ht]
    \centering
    \includegraphics[width=1\linewidth]{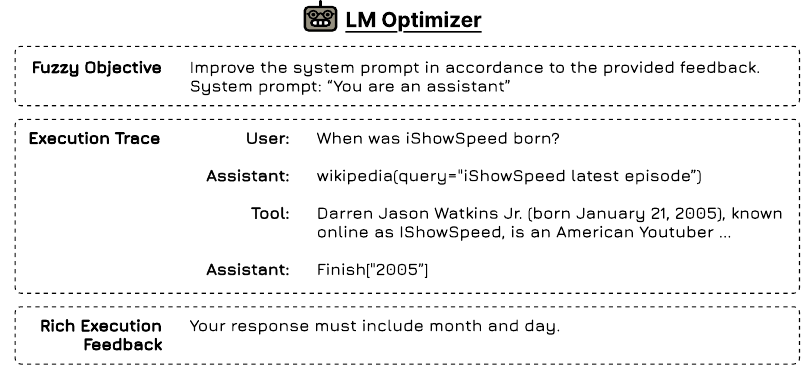}
    \caption{\textbf{LM optimizers interpret fuzzy objectives and execution trace with rich execution feedback}. We illustrate a simple example with an agent. In practice, the objective contains the variables to optimize, and the execution trace shows how these variables were used. Optionally, execution feedback can be provided.}
    \label{fig:optimizerexample}
\end{figure}

\begin{figure}[ht]
    \centering
    \includegraphics[width=0.75\linewidth]{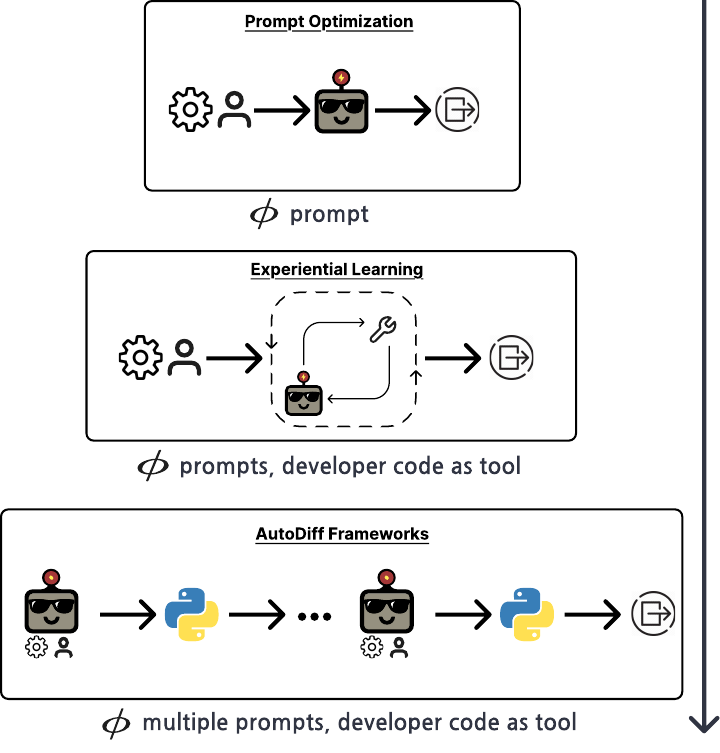}
    \caption[\textbf{Evolution of LM optimization approaches}.]{\textcolor{changemark}{\textbf{Evolution of LM optimization approaches}. LM optimization approaches have evolved from prompt optimization, then agents, and finally workflows. In all approaches, an LM acts as an optimizer for generating updates on prompts or tools. \textbf{Top:} prompt optimization focuses on LMs as opposed to scaffolded LMs. The LM optimizer uses a single or multiple LM calls. \textbf{Middle:} experiential learning extends to agents and focuses on extracting insight from tasks and storing them in the system message, or refining tools. The LM optimizer uses a single LM call to update them from successful or unsuccessful execution traces. \textbf{Bottom:} AutoDiff frameworks focus on workflows where computation is represented as a graph interleaving LM calls and developer-defined code as tools. Using textual feedback, the LM optimizer can optimize nodes of the graph (e.g., system prompts).}}
    \label{fig:optimizers}
\end{figure}

{\color{changemark}We view scaffolded LM $\pi$ as semi-parametric models where the LM is the parametric component that is fixed.
Instead, we consider the training of non-parametric variables, prompt and tools, that are represented in language and code.
As shown in Fig. \ref{fig:optimizers}, non-parametric variables include the prompt and developer-defined code as tools.
As shown in Fig. \ref{fig:optimizerexample}, we use an LM as an optimizer to generate the updated prompt or tool according to an objective (``improve prompt according to feedback on this execution trace''), an execution trace, and optionally execution feedback $f$, all provided in natural language.
}

As in machine learning (ML), we assume a training, validation, and test dataset sampled from the same (unknown) distribution.
{\color{changemark}During training, we iteratively sample data point(s) and update prompt or tool.
At inference, prompt(s) and tools are fixed (in the case of workflows there's a multiple prompts, i.e., a different prompt for each step).
Unlike in ML, where we train with Monte Carlo samples to prevent catastrophic forgetting \cite{Hadsell2020}, an LM optimizer works with variables represented in language or code as opposed to tensors.
Thus, the role of batching is still an open research question \cite{textgrad}.
The datasets $\mathcal{D}$ consist of user message corresponding to different tasks from which we obtain execution traces $\tau \sim \pi$.
During training, the execution traces come from any scaffolded LM $\pi^{\prime}$ or the current scaffolded LM $\pi$.
In that sense, it shares similarities with RL, where we have on/off-policy training and offline/online training \cite{Sutton2018}.
}%
However, the training dynamics significantly differ from those in RL.
While RL is about finding good data (exploration) and imitating that good data (exploitation) \cite{eysenbach2020supervised}, recent works in Sec. \ref{sec:autodiff} learn only from bad execution traces.
This is because failed execution traces often yield rich textual feedback, such as error messages or simulated user responses.

Central to this paradigm is an optimizer implemented with an LM as opposed to bootstrapping successful execution traces as in-context learning examples \cite{dsp, learnincontext}.
{\color{changemark}We further discuss how this paradigm differs from other approaches that have employed an LM and non-parametric components for improving a scaffolded LM.
}%
First, in Reflexion \cite{reflexion}, the updates are not retrained across tasks and are only used for retrying a task.
Second, CoALA \cite{sumers2024cognitive} is an orthogonal approach that works at run-time with a specific scaffold.
In particular, it requires changing the design of the agent with a memory module.
{\color{changemark}This memory module not only serves as long-context memory but also for improving the agent over time.
}%
However, this requires the agent to make additional decisions on what and when to store or retrieve knowledge from this memory module.
In training with language supervision, recent works focus on a setting where we are provided with a fixed dataset for training.
{\color{changemark}Thus, it can be seen as amortizing the expensive inference of a cognitive architecture \cite{sumers2024cognitive} into a training procedure.
}

Our organization follows the structure outlined in Fig. \ref{fig:optimizers}.
Tracing their development, these methods have evolved to optimize more complex execution traces.
Automatic prompt optimization focuses on an LM (as opposed to a scaffolded LM).
It covers various types of objectives for improving the prompt of a target LM.
Experiential learning trains agents where the tools and system prompts matter.
Accordingly, the execution traces consist of interactions between the LM and the tools, from which the LM optimizer extracts reusable tools or insights.
Finally, AutoDiff frameworks focus on workflows, which resemble classical software pipelines that include LM calls at key stages.
These frameworks focus on recording the execution trace as a graph.
The LM optimizer then uses this graph for execution and optimization.

\subsection{Prompt Optimization}
\label{sec:promptoptim}

{\color{changemark}In prompt optimization \cite{li2025surveyautomaticpromptengineering}, we update the prompt of an LM (as opposed to a scaffolded LM).
We prompt an LM optimizer with an objective (e.g., ``improve this instruction according to the following examples of tasks where the model fails'') and (optionally) a textual feedback $f$.
The updates made by the LM optimizer is evaluated on a validation set and the best one is selected.
Thus, while these methods use a different objective in natural language, by heuristically selecting the best update on a validation set, they implicitly optimize for performance on the validation set.
}
In its simplest form, we provide query and response examples and prompt the LM to generate the corresponding instruction \cite{dspy, ape}.

Provided with different initial instructions, other methods define the objective as generating variations.
Each variation is evaluated on a validation set, and the top-performing versions are retained.
In \cite{ape}, the objective is to generate semantically similar instructions.
Evolutionary methods \cite{promptbreeder, evoprompt} combine parts of multiple instructions and then introduce random changes; we call this process crossover and mutation.
In OPRO \cite{opro}, we define the objective as generating instructions with a higher score using previous instructions, their scores, and a random sample of input-output pairs from the training data.
To introduce fine-grained variations of instructions, SAMMO \cite{sammo} uses a more granular representation of prompts by breaking them down into smaller sections, such as input formatting.
This fine-grained structure allows SAMMO to adjust specific components, such as changing the input format from raw text to a structured format (e.g., XML).
In a similar vein, ABO \cite{abo} implements the instruction as step-by-step guidelines.
Thus, the LM optimizer only updates the specific step that failed.

Another line of work uses an LM to generate rich textual feedback $f$, and defines the objective as following the feedback $f$ \cite{protegi, promptagent, reprompt, gradsum}.
Importantly, textual feedback is obtained from sample queries where the target LM fails.
Additionally, these works also propose summarizing multiple pieces of feedback \cite{reprompt} or updates \cite{gradsum}.

Self-discover \cite{self-discover} focuses on the output structure of the instruction.
The objective is to generate a JSON format with keys representing distinct reasoning steps.
By filling in each key in an autoregressive manner during generation, it guides the target LM's reasoning steps.

\subsection{Experiential Learning}
\label{sec:expel}

\begin{figure}[ht]
    \centering
    \vspace{7pt}
    \includegraphics[width=\linewidth]{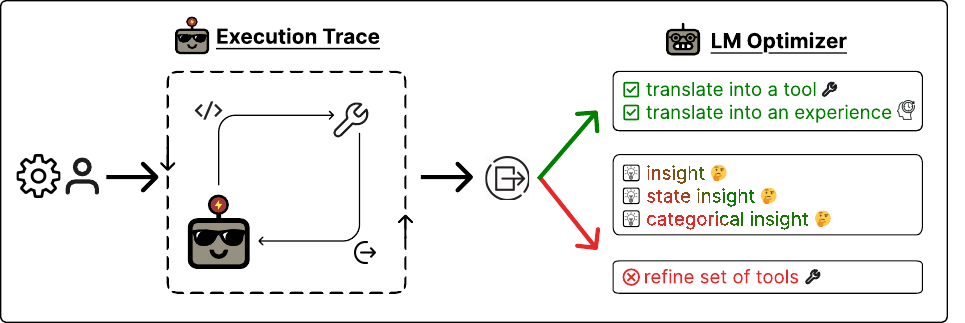}
    \caption[\textbf{Experiential learning focuses on tools and insights.}]{\textbf{Experiential learning focuses on tools and insights.} Successful execution traces are used for extracting reusable tools \cite{wang2025inducingprogrammaticskillsagentic} while unsuccessful execution traces are used to refine tools \cite{agentoptimizer}. Additionally, we can contrast successful and unsuccessful trajectories on the same task to extract insights \cite{zhao2024expel}.}
    \label{fig:expel}
\end{figure}

We borrow the term experiential learning from ExpeL \cite{zhao2024expel} to refer to the training of agents with language supervision.
{\color{changemark}The LM optimizer considers an objective (e.g., ``extract a general insight on why you failed this task'', ``extract reusable tool from execution trace''), prompt $\mathbf{s}_0$, execution trace $\tau$, optionally a textual feedback $f$, and generates an insight or a tool.
In particular, the execution trace $\tau$ of agents consists of a sequence of tool calls/responses and ends with a response to the user.
}
As discussed in Sec. \ref{sec:agent}, tools play an important role in agents.
{\color{changemark}As shown in Fig. \ref{fig:expel}, the objective corresponds to either improving the set of tools or extract insights that are provided to the agent's system prompt.
}%
Specifically, prior works leverage successful and unsuccessful execution traces with different objectives.

\subsubsection{Insights Extraction}
\label{sec:insight}

{\color{changemark}ExpeL \cite{zhao2024expel} starts by collecting a dataset for training consisting of prompt $\mathbf{s}_0$ and corresponding execution trace $\tau$.
Using a Reflexion agent \cite{reflexion}, multiple execution traces may be sampled for a prompt $\mathbf{s}_0$.
In particular, the Reflexion agent allows the agent to retry the task upon failing, which allows the collection of unsuccessful and successful trajectories $(\tau^\text{succ}, \tau^\text{fail})$ on the same prompt instance $\mathbf{s}_0$.
The LM optimizer refines a list of insights from successful and unsuccessful execution traces $(\tau^\text{succ}, \tau^\text{fail})$ for a given prompt $\mathbf{s}_0$ or from $k$ successful trajectories $(\tau_1^\text{succ}, ..., \tau_k^\text{succ})$.
}%
At inference, those insights are fixed and provided to the scaffolded LM through its system prompt.

Follow-up works on ExpeL \cite{zhao2024expel}, such as AutoGuide \cite{fu2024autoguide}, modify insights to be selectively added to the context at each step.
Insights are specific to a state and contain a description of the state in which it is applicable.
%
%
However, these methods require additional cost at inference since an LM summarizes the current state and then uses it to select the corresponding state-specific insight.
AutoManual \cite{automanual} modifies ExpeL \cite{zhao2024expel} to prevent distributional shift in offline RL, where at test-time the state distribution is different \cite{offlinetoonline}.
However, their paper does not show strong evidence that training agents from language supervision suffers from distributional shift \cite{offlinetoonline}.
Additionally, MSI \cite{Fu2024MSIAgentIM} proposes to categorize each insight and retrieve the relevant one at test-time based on pre-defined rules.

{\color{changemark}%
ExpeL \cite{zhao2024expel} also explores new settings such as transfer learning, where we use insights learned from a source dataset $\{\mathbf{s}_0^{\text{source}}\}$ and transfer it to a target dataset $\{\mathbf{s}_0^{\text{target}}\}$.
}%
In parametric training \cite{pouyanfar2018survey}, it consists of initializing the weights with the model trained on the source dataset.
Then, finetuning these weights on the target dataset.
In ExpeL, they initialize with insights extracted from a source dataset.
Then, the LM optimizer's objective is to generate refined insights to the target task provided with an execution trace (from a human).
Note that the update is done in a single shot thanks to the LM's ability to interpret intricate objectives.
Additionally, they manually inspect execution traces from the agent with and without the insights.
For instance, in ALFWorld \cite{Shridhar2020ALFWorldAT}, one insight would be ``when searching for an item, consider its nature and its typical usage.''
This leads the agent to update its belief \cite{andreas-2022-language} and avoid unnecessary actions to find objects.
Furthermore, they find that another insight, ``if an attempt to interact with an item fails [...] consider alternative actions or locations,'' leads to self-correction.
{\color{changemark}These findings are important, they showcase that simple \emph{high-level insights influence the behavior of agents in some predictable way}.
}

\subsubsection{Tools Optimization}
\label{sec:toolopt}

Tool optimization \cite{agentoptimizer, learnact, wang2025inducingprogrammaticskillsagentic} focuses on refining developer-defined code into more effective routines.
The main challenge of tool optimization is ensuring that the skills are reusable and useful across episodes \cite{wang2024voyager}.

A related term to tool, but slightly more general is skill \cite{wang2024agentworkflowmemory}.
It refers to both a textual representation of a sequence of action and developer-defined code as tool.
The term ``skills'' emphasizes two aspects of tool optimization.
First, we want to make tools more efficient, thereby achieving tasks with fewer steps.
Second, similarly to unsupervised RL (URL) \cite{moss, eysenbach2018diversity}, we want to extract the set of behaviors possible in the environment.
Drawing an analogy with principal component analysis, before we are given any tasks, we want to know what is the basis of behavior that exists in some environment.
Similarly, these methods \cite{liu2025contextual, wang2024agentworkflowmemory} stores skills into a vector database and retrieve corresponding skill at inference.
For instance, in CER \cite{liu2025contextual} skills are represented as a sequence of tool calls with a high-level description of the skill.
In AWM \cite{wang2024agentworkflowmemory}, they explore different textual representation of skills.
We can also understand these methods as `curating' in-context samples for retrieval of similar experiences in ExpeL \cite{zhao2024expel}.
However, the textual representation represents a challenge for formal verification \cite{wang2025inducingprogrammaticskillsagentic}.

{\color{changemark}At each step, LearnAct \cite{learnact} and AgentOptimizer \cite{agentoptimizer} sample an execution trace $\tau \sim \pi(\mathbf{s}_0)$, and refine the set of tools for unsuccessful execution traces $\tau^{\text{fail}}$.
LearnAct uses execution feedback $f$ corresponding to execution errors.
}%
The LM optimizer's objective is to address $f$ by either revising, adding a tool, or updating the system prompt with information about the tools.
AgentOptimizer does not rely on the execution feedback $f$.
The LM optimizer modifies the set of tools at each step by revising, removing, or adding a tool to solve the task that the model failed on.
Moreover, they propose heuristics to account for the LM's inability to correctly understand the failure modes and update tools.
At each epoch, they evaluate the performance of the agent on the training data.
If the performance is lower than the previous epoch, they roll back the set of tools.
However, this is expensive and quickly leads to overfitting, where tools are not reusable.

In contrast, ASI \cite{wang2025inducingprogrammaticskillsagentic} extracts reusable tools from a successful execution trace.
One important challenge in extracting a reusable tool is to make sure that it is both useful and bug-free.
AgentOptimizer \cite{agentoptimizer} checks the performance on the training set with and without the updated tools by simulating again on the same query.
An LM is used to rewrite the execution trace with only the extracted tools.
Then, the agent attempts to finish the execution.
{\color{changemark}If it successfully solves the task, then the extracted tool is kept.
}%
In contrast to transfer learning for insights, ASI explores domain adaptation \cite{ovadia2019trustmodelsuncertaintyevaluating} in web browsing.
They find that the agent can effectively reuse tools and adapt tools to the new domain.

Similar to unsupervised RL \cite{moss}, recent works propose to use an LM to both propose tasks and extract tools from rollouts on these tasks.
In Voyager \cite{wang2024voyager}, the LM optimizer refines tools based on execution feedback and adds them to the skill set when successfully used to achieve a task.
{\color{changemark}SkillWeaver \cite{zheng2025skillweaverwebagentsselfimprove} extracts tools directly from successful execution traces.
}%
Then, it produces test cases to check the long-term viability of the tool.

\subsection{AutoDiff Frameworks}
\label{sec:autodiff}

The term autodiff is inspired by deep learning frameworks \cite{paszke2019pytorch} in the sense that it implements elementary operations for defining and optimizing workflows \cite{textgrad, trace, symboliclearning, adalflow}.
The frameworks represent execution traces with a directed acyclic graph (DAG) where nodes correspond to variables or operations and edges represent dependencies.
{\color{changemark}They roll out execution traces $\tau\sim \pi$ and obtain execution feedback $f$.
}%
Subsequently, prior works \cite{textgrad, adalflow, symboliclearning, wang2025how} use the chain rule to back-propagate $f$ through the graph.
Each variable of the graph is updated by the LM optimizer according to its feedback.
Alternatively, Trace \cite{trace} converts the DAG into text akin to a Python traceback.
The LM optimizer updates variables in an autoregressive manner.
We organize this section as follows.
First, we cover graph abstraction to understand which variables of the workflow are optimized in different frameworks.
Second, we focus on how execution traces are recorded and used for optimization.

\subsubsection{Graph Abstraction}
\label{sec:graphabstraction}

We refer to graph abstraction as the representation of nodes in the graph.
This abstraction is foundational because it determines how workflows are formalized, traced, and subsequently optimized.
There are two types of nodes in the graph: variables and operations.
Variables correspond to non-parametric components of the workflow (e.g., system prompt) or intermediate values (result from operations).
{\color{changemark}Operations correspond to computation done on variables, e.g., Python functions, LM calls.
}%
As an analogy with graph representation for neural networks, variables would be weights, and operations would be layers.
However, different from neural networks, any node of the computational graph can be optimized, including operations \cite{trace}.
We distinguish between two lines of work, focusing on prompts or Python functions.
It is important to note that in this survey, we focus on workflows within the scope of scaffolded LM.
Thus, we do not consider applications of these frameworks for optimization instances (as opposed to workflows).

TextGrad \cite{textgrad} and other variants \cite{adalflow, wang2025how, symboliclearning} focus on textual variables.
In particular, operations include concatenations and LM calls \cite{textgrad}.
Accordingly, trainable variables correspond to the system prompt or user input template of an LM call.

Trace \cite{trace} focuses on general-purpose workflows.
Importantly, operations are not limited to LM calls, they can be arbitrary Python functions.
In practice, we wrap Python functions with the Python decorator ``bundle''.
Additionally, they support the optimization of nodes corresponding to operations.
{\color{changemark}In that sense, it goes beyond prompts to optimizing the workflow's developer-defined tool.
}
\begin{figure}[ht]
  \centering
  \begin{minipage}{0.48\textwidth}
    \begin{prettypython}[title=Chain Rule]
# step 1
output = workflow(user_message)
feedback = compute_feedback(output)

# step 2, compute the grad field of variables
feedback.backward()

# step 3, update each variable according to grad
optimizer.step()
    \end{prettypython}
  \end{minipage}
  \hfill
  \begin{minipage}{0.48\textwidth}
    \begin{prettypython}[title=Formatting of execution trace]
# step 1
output = workflow(user_message)
feedback = compute_feedback(output)

# step 2, format prompt template of optimizer
output.backward(feedback)

# step 3, update variables autoregressively
optimizer.step()
    \end{prettypython}
  \end{minipage}
    \caption[\textbf{Graph execution and optimization using chain rule vs. formatting.}]{\textbf{Graph execution and optimization using chain rule \textit{vs.} formatting of execution trace}. AutoDiff frameworks consist of three key steps: (1) forward computation, (2) backward feedback propagation, (3) variable updates. Both approaches differ during their backward pass. \textbf{Left:} using the chain rule, we compute a grad field on each variable corresponding to a feedback that addresses the feedback on its child variable. The LM optimizer updates each trainable variable following its grad field. \textbf{Right:} it performs a reverse topological sort on the computation graph starting from the output to format the prompt template of the optimizer. Together with the feedback, the LM optimizer generates trainable variables in an autoregressive manner.}
    \label{fig:graphexec}
\end{figure}

\subsubsection{Graph Execution and Optimization}
\label{sec:trace}

\begin{figure}[ht]
\centering
\vspace{7pt}
\scalebox{0.85}{ 

\begin{tikzpicture}[
    param/.style={draw, rounded corners, fill=mygreen!75, minimum width=2.8cm, minimum height=0.8cm, text centered},
    var/.style={draw, rounded corners, fill=blue!10, minimum width=2.8cm, minimum height=0.8cm, text centered},
    op/.style={draw, rounded corners, fill=harvestgold!70, minimum width=3.5cm, minimum height=0.8cm, text centered},
    fwd_arrow/.style={-{Stealth[scale=1.2]}, thick, black}, 
    bwd_arrow/.style={-{Stealth[scale=1.2]}, thick, red, dashed}, 
    arrowlabel/.style={font=\footnotesize\it},
    circlenode/.style={circle, draw, fill=white, text=red, font=\small, minimum size=0.6cm} 
]

{\color{changemark}\node[param] (prompt) at (1,0) {system message};
\node[param] (data) at (7,0) {user message};
\node[op] (concat) at (5,-2) {concat(system message, user message)};
}\node[param] (input) at (5,-3.5) {input};
\node[op, minimum width=4.5cm] (lm) at (5,-5) {LM call};
\node[param, minimum width=3.8cm] (output) at (5,-6.5) {response};
\node[op, minimum width=4.5cm] (score) at (5,-8) {evaluate(output)};
\node[param, minimum width=4.5cm] (feedback) at (5,-9.5) {feedback};

\draw[fwd_arrow] (prompt) -- (concat);
\draw[fwd_arrow] (data) -- (concat);
\draw[fwd_arrow] (concat) -- (input);
\draw[fwd_arrow] (input) -- (lm);
\draw[fwd_arrow] (lm) -- (output);
\draw[fwd_arrow] (output) -- (score);
\draw[fwd_arrow] (score) -- (feedback);

\draw[bwd_arrow, bend left=15] (feedback) to (output);
\draw[bwd_arrow, bend left=15] (output) to (input);
\draw[bwd_arrow, bend left=15] (input) to (prompt);


\end{tikzpicture}
}
\caption{\textbf{Chain rule in TextGrad.} Forward computation is represented with solid black arrows. Calling ``feedback.backward()'' (dashed red arrows, bent) applies the chain rule on the feedback $f$. It computes a gradient for each variable (in green), provided with gradient functions attached to operations (in yellow).}
\label{fig:textgrad}
\end{figure}

{\color{changemark}We refer to graph execution and optimization as the process of sampling an execution trace $\tau\sim \pi(\mathbf{s}_0)$ and optimizing variables of the graph according to the execution feedback $f$.
}%
As shown in Fig. \ref{fig:graphexec}, it involves three key steps: (1) forward computation, (2) backward feedback propagation, (3) variables update.
{\color{changemark}During forward computation, we collect execution feedback $f$.
}%
Then, backward feedback propagation performs a reverse topological sort to either apply the chain rule on the execution feedback \cite{textgrad, adalflow, wang2025how, symboliclearning} or format a prompt template with the DAG \cite{trace}.
On the one hand, the chain rule enables structured credit assignment by propagating feedback through the graph in a fine-grained manner.
On the other hand, encoding the entire execution trace as a single prompt template is more token-efficient.

In the context of optimization in the language space, the chain rule does not refer exactly to the differentiation of composite functions in calculus.
Instead, it points to the idea of when a transformation is applied to the output of another transformation.
Then, the overall effect on an input can be understood by systematically combining the influence of each transformation in sequence.
{\color{changemark}As shown in Fig. \ref{fig:textgrad}, and following the graph abstraction presented in Sec. \ref{sec:graphabstraction}, operations, including LM calls and concatenation, create nodes and attach a grad function.
}%
Most importantly, the grad function contains a prompt for an LM that informs about that operation's role.
Similarly, each variable has a role.
These roles are provided in text to account for heterogeneous variables and operations in workflows.
Accordingly, the backward feedback propagation calls the attached grad function at each operation in the reverse order of the DAG.
Each gradient is a textual feedback with respect to the child node of that operation.
For each trainable variable, the LM optimizer's objective follows the feedback as well as the variable's role.
Follow-up work extends Textgrad \cite{textgrad} in the following ways.
AdalFlow \cite{adalflow} extends operations to retrievers \cite{rag}.
{\color{changemark}GASO \cite{wang2025how} modifies the gradient operation to take into account neighboring nodes.
}

\begin{figure}[ht]
    \centering
    \vspace{7pt}
    \includegraphics[width=1\linewidth]{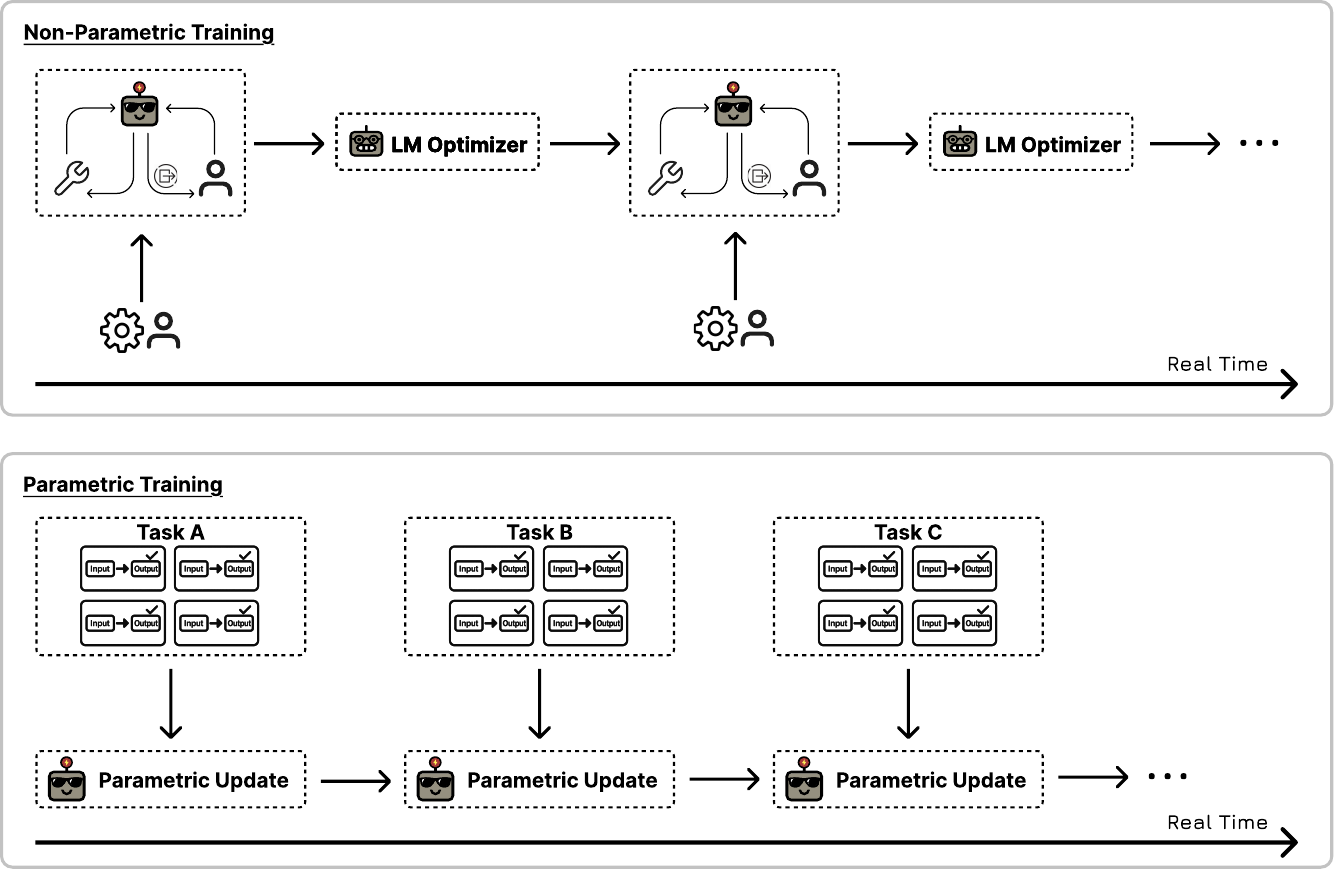}
    \caption[\textbf{Continual learning with non-parametric vs. parametric training.}]{\textcolor{changemark}{\textbf{Continual learning with non-parametric \textit{vs.} parametric training.} \textbf{Top:} we focus on non-parametric training where users provide rich textual feedback at each episode. This feedback is naturally present as users interact with the agent to adjust the answer. We update the agent with an LM optimizer after each episode, allowing agents to learn across episodes \cite{silver2025era}. \textbf{Bottom:} parametric training uses scalar feedback and focuses on solving catastrophic forgetting that occurs when the \textit{i.i.d} assumption doesn't hold at each gradient step \cite{Hadsell2020}.}}
    \label{fig:iid}
\end{figure}

A conceptually more efficient approach is Trace \cite{trace}.
At each step of the optimization process, the chain rule requires multiple calls to an LM.
Furthermore, the LM optimizer also requires multiple calls to an LM.
In contrast, Trace \cite{trace} formats a prompt template for the execution trace during backward feedback propagation.
This means the LM optimizer generates all updated variables in an autoregressive manner, provided with the execution trace and feedback.
Additionally, as discussed in Sec. \ref{sec:graphabstraction}, Trace \cite{trace} uses a different graph abstraction that allows for arbitrary Python functions as operations.
As opposed to Textgrad \cite{textgrad}, this allows the LM optimizer to also consider operations during training.
{\color{changemark}In that sense, we can optimize the workflow's developer-defined tool.

\section{Beyond episodic learning}
\label{sec:beyondiid}

This section discusses non-parametric training as an approach towards \emph{agents that inhabit streams of experience}\footnote{We purposely focus on agents, as mixed autonomy settings typically involve an agent \cite{silver2025era, yao2025second}.}.
%
We describe it as agents that continuously learn from experience.
In particular, current models only learn and improve from their mistakes within an episode.
At the end of the episode, the chat history is cleared and the model will recommit the same errors and rediscover the same solutions in future episodes.

This key capability of learning beyond episodic interaction is crucial for mixed-autonomy settings \cite{cursor2023, microsoft2025copilot}.
In this setting, there's significant contextual recurrence across episodes; tasks exhibit temporal coherence, wherein humans and AI dynamically share control based on their respective strengths.
It also features rich and interpretable feedback, where humans identify errors and offer guidance \cite{tau-bench,zhou2025sweetrltrainingmultiturnllm}.
For instance, agents rarely succeed on the first attempt, as user prompts often omit crucial information—highlighting a mismatch between latent intent and explicit input \cite{van-duijn-etal-2023-theory}.
As a result, each episode may involve multiple rounds of user feedback, where the agent corrects its behavior.
As illustrated in Fig. \ref{fig:iid}, we propose a setting in continual learning with non-parametric updates as opposed to parametric updates.
Unlike scalar-based update, language-based update is interpretable and does not suffer from catastrophic forgetting \cite{Hadsell2020}, while being compatible with closed-source models.

Below, we discuss key capabilities required for agents to inhabit streams of experience using non-parametric training with language supervision.
To support such learning, we use an LM as an optimizer that interprets rich user feedback, execution traces, and intricate objectives (e.g., ``minimize redundant steps based on prior interaction history'', ``align with user-specified constraints''), and update non-parametric variables.
Compared to parametric training, it uses rich textual feedback and does not suffer from sparse rewards \cite{creditassignment} and catastrophic forgetting \cite{Hadsell2020}.
Each episode begins when the human delegates control to the AI via a user query.
The AI then autonomously interacts with tools to attempt task completion.
After producing an initial result, the human may intervene to offer feedback—correcting errors and providing guidance.
The AI integrates this feedback and continues the task.
The episode concludes once the AI successfully completes the task or the human resumes control to begin a new one.
In formula, given the system message and user message $\mathbf{s}_0$, an episode $\tau$ is a sequence of LM's response $\mathbf{a}_t$, tool response $\mathbf{o}_t$, and human feedback $f$
$$
\tau = ( \mathbf{s}_0, \mathbf{a}_0, \mathbf{o}_0, ..., f, ..., \mathbf{a}_t, \mathbf{o}_t, ..., f, ..., \mathbf{a}_{T}).
$$
At the end of each episode, a non-parametric update is applied to the scaffolded LM using an LM optimizer.
We posit that an effective learning system in this setting must exhibit several key properties:
}\begin{itemize}
{\color{changemark}    \item \textbf{Scalable learning across similar tasks.} The system should improve as it accumulates experience. This defines a new scaling axis—performance gains through recurrence. For instance, maintainers resolve GitHub issues 5-18x faster than external contributors \cite{kwa2025measuringaiabilitycomplete}, suggesting the value of task familiarity.
    \item \textbf{Plasticity and stability.} The model must quickly adapt to new tasks (plasticity) while not forgetting how to solve tasks it previously learned (stability). If it forgets, it should recover rapidly from supervision.
    \item \textbf{Efficiency.} It should avoid excessive context consumption to remain effective over long-term interactions spanning weeks to months.
    \item \textbf{Interpretability.} Especially in mixed-autonomy settings, the AI's internal state must remain legible so that humans can anticipate its behavior. Carrying over entire trajectories verbatim hinders interpretability.
}\end{itemize}

{\color{changemark}

\subsection{Scalability}
\label{subsec:scalability}

%
In mixed-autonomy deployments, users repeatedly face related sub-problems under shared tools, codebases, or data silos.
This induces \emph{recurrence}: the same APIs, preconditions, and failure modes reappear across episodes.
If a scaffolded LM fails to improve under recurrence, it repeatedly pays exploration cost and re-commits prior mistakes.
We therefore target scaling for tasks that share similar context (e.g., same codebase): systematic gains as similar tasks recur, achieved by updating non-parametric variables (prompts, tools, step policies) with \emph{language supervision} rather than gradient steps.

Streams of experience naturally provide \emph{textual} signals—human corrections, rationales, constraint reminders, and execution errors—alongside verifiers (unit tests, checkers, compilers) when available.
Such signals (i) localize error causes (wrong tool, argument, or precondition), (ii) support immediate, auditable edits to prompts and tools, and (iii) avoid catastrophic forgetting inherent to weight updates \cite{Hadsell2020}.

We highlight two mechanism for scaling in this context:
\begin{itemize}[leftmargin=1.4em,itemsep=2pt,topsep=3pt]
  \item Insight caching (Sec.~\ref{sec:insight}): distill state-invariant heuristics or decision rules from traces (e.g., ``if login required, navigate to \texttt{/signin} before form submit'') and insert them into high-privilege prompts \cite{zhao2024expel, fu2024autoguide}.
  \item Skill/tool induction (Sec.~\ref{sec:toolopt}): refactor recurring action patterns into reusable, typed tools with pre/post-conditions and minimal tests \cite{wang2025inducingprogrammaticskillsagentic, learnact, agentoptimizer, zheng2025skillweaverwebagentsselfimprove, wang2024voyager}.
\end{itemize}

In-episode reflection \cite{reflexion} improves retries but does \emph{not} persist if no artifact is updated. RL-tuned reflection controllers (e.g., better verbal feedback and plans) still pay the same exploration cost when a similar task reappears, unless insights or tools are written back \cite{retroformer}.
In contrast, \emph{agents that inhabit streams of experience} synthesize durable artifacts from traces: principle extraction from failures, explicit tool learning.

Prior work equips agents with a memory module \cite{generative-agents}.
This has long been seen as an important step towards self-evolving agents \cite{agentsurvey, openai2025memory}.
For instance, in ExpeL \cite{zhao2024expel}, they save successful trajectories in an experience buffer.
At inference, it retrieves similar experiences and appends them to the context of the agent.
Similarly, Dynamic Cheatsheet \cite{suzgun2025dynamiccheatsheettesttimelearning} stores and retrieves accumulated strategies, code snippets, and general problem-solving insights at inference time.
While this helps the agent's performance, it does not scale well.
An important property in the recent breakthrough of AI is the scaling with more data points \cite{sutton2019bitter}.
We expect the agent to find more efficient solutions as it encounters similar tasks.
Retrieving similar experiences is limited to a few experiences.
Generative agents \cite{generative-agents} use the memory module to simulate real human behaviors that plan and act over very long horizons.
In particular, the focus of generative agents is to account for the limited context of LMs by ``compressing'' previous interactions and retrieving them later.
However, these works focus on simply storing experiences \cite{zhao2024expel} or compressing prior interactions \cite{generative-agents} to fit multiple interactions beyond the context window.



\subsection{Plasticity and Stability}
\label{subsec:robustness}

Agents that inhabit streams of experience must \emph{adapt quickly} (plasticity) while \emph{not regressing} on previously mastered patterns (stability).
Language-level edits (prompts, tools, validators) help by localizing changes, preserving provenance, and enabling rollback—circumventing catastrophic forgetting common in parametric updates \cite{Hadsell2020}.

Plasticity and stability requires tracking beyond binary success \cite{streambench,llmevolve}:
\begin{itemize}[leftmargin=1.4em,itemsep=2pt,topsep=3pt]
  \item Assistance rate: mean human interventions per task. We want to reduce the assistance rate over time without affecting the assistance rate of previous tasks.
  \item Backward/forward transfer (BWT/FWT): BWT measures accuracy on prior tasks after an edit; FWT measures improvement on new tasks after learning on previous tasks.
\end{itemize}
As an analogy, we want to implement new features to a codebase without breaking previous features \cite{jimenez2024swebench}. We highlight some mechanism to preserve stability:
\begin{itemize}[leftmargin=1.4em,itemsep=2pt,topsep=3pt]
  \item Scoped edits: restrict updates to the smallest locus (one tool, one step prompt) via graph credit assignment \cite{textgrad, trace}.
  \item Gated promotion: accepte update only if they pass unit tests \cite{jimenez2024swebench}.
  \item Versioned artifacts: keep prior versions + provenance (trace, feedback, objective) for audit and instant rollback \cite{wang2025inducingprogrammaticskillsagentic, agentoptimizer}.
\end{itemize}

\subsection{Efficiency}
\label{subsec:efficiency}

Dumping prior trajectories or long memories into context windows scales poorly as context-windows are limited.
Non-parametric training instead \emph{compresses experience into durable artifacts}—concise insights, executable tools, and step-specific prompts—so future episodes consume similar or fewer tokens while solving harder variants. For instance, we can update prompt and tools with
\begin{itemize}[leftmargin=1.4em,itemsep=3pt,topsep=3pt]
    \item priors: encode stable facts about the environment: API maps, repo layout, naming conventions, data schemas, units, and invariants. This allows fewer exploratory calls.
    \item constraints: specifies constraints according to user preferences.
    \item docstring: short, copy-ready examples and counter-examples in the tool description.
\end{itemize}

Besides considering the efficiency of the scaffolded LMs, it is important to also consider the LLM optimizer's efficiency.
Inspired by sleep-time compute \cite{lin2025sleeptimecomputeinferencescaling}, we can imagine separating \emph{online} learning from \emph{offline} consolidation to preserve latency.
In particular, recent work proposed the concept of sleep-time compute \cite{lin2025sleeptimecomputeinferencescaling}.
This paper is motivated by the fact that LM user queries consist of a shared task context $c$ and queries.
Thus, when the model is idle, it proactively updates the shared context by inferring likely future queries and restructures the context in ways that may be beneficial at test time.

\subsection{Interpretability}
\label{subsec:interpretability}

In mixed autonomy, humans must anticipate and steer agent behavior. Carrying opaque internal states or long, free-form memories across episodes hinders oversight.
Non-parametric updates promote \emph{legibility}: each change is a human-readable diff, on prompt and tools, linked to the trace and feedback that motivated it.
In particular, working with language and code allows to easily track:
\begin{itemize}[leftmargin=1.4em,itemsep=2pt,topsep=3pt]
  \item Provenance: instead of storing tensors, it stores pure text.
  \item Diffs: relative changes that happen at each step are easily interpretable.
  \item Information: the hierarchical structure of the chat format allows to control how information is interpreted \cite{wallace2024instructionhierarchytrainingllms}.
\end{itemize}

\noindent Additionally, several effective approaches operate on the \emph{parametric} component, offering adaptation but typically with weaker legibility than prompt/tool edits:


\paragraph{Continual learning.}
In parametric training, continual learning aims to address a core limitation of parametric training: catastrophic forgetting.
Notably, models are optimized using \textit{i.i.d.} batches sampled from a fixed dataset.
}%
However, consider a setting where a model is first trained to classify dogs, and later needs to be extended to also classify cats.
Simply continuing training on the new data can lead to the model overwriting its knowledge of dogs—resulting in a phenomenon known as catastrophic forgetting.
A naive solution might involve freezing the feature extractor and training only the output head.
Yet this often leads to poor performance, as the learned features are not well-suited for the new task (e.g., distinguishing cats).
{\color{changemark}Continual learning addresses this by developing methods that preserve previously learned knowledge while integrating new information, without requiring retraining on the full original dataset.
}%
Its motivation is different: to avoid the inefficiency of reprocessing previously seen data—especially when that data has already been well-classified.
The key insight is that gradient-based learning on i.i.d. samples can be wasteful when the model only needs to adapt at the margin.
Furthermore, gradient-based learning faces the plasticity-stability dilemma \cite{nikishin2022primacybiasdeepreinforcement, achille2019criticallearningperiodsdeep, frankle2020earlyphaseneuralnetwork, elsayed2024streamingdeepreinforcementlearning}: models are highly plastic early in training, quickly adapting to new data, but the plasticity decreases over time.
{\color{changemark}As shown in Fig. \ref{fig:iid}, besides the aforementioned challenges, continual learning would also require humans to provide labels (in the form of reward or demonstration).
In contrast, updating non-parametric variables does not exhibit a decaying plasticity over time and mixed-autonomy settings naturally provides feedback in the form of textual corrections and guidance.


%

\paragraph{Test-time training.}
test-time training methods \cite{karami2025latticelearningefficientlycompress} function as memory mechanisms that compress information directly within the embedding space, rather than relying on text-based representations.
}%
For instance, Lattice \cite{karami2025latticelearningefficientlycompress} introduces a compression model that dynamically updates and maintains a compact representation of contextual history by operating on the key-value caches of a transformer in a streaming fashion.
These methods typically involve architectural modifications to the neural network and offer an orthogonal approach to standard prompting or finetuning.
Potentially, the right objective function for these methods could enable emergent behaviors—such as learning across episodes—in the embedding space.

{\color{changemark}

}
\section{Conclusion}

This survey introduces a paradigm called training of scaffolded LMs with language supervision.
It organizes the intricate literature in prompt optimization, LM pipelines, experiential learning agents, AI workflow optimization, and LM agents.
We focus on the structure around LMs where AI and human share control and decision-making responsibilities to achieve a common goal \cite{openai_charter, openhands, windsurf_editor}, and refer to this structure as the scaffold.
We view scaffolded LMs as semi-parametric models where the scaffold represents the non-parametric component and the LM refers to the parametric component.
From a learning perspective, by focusing on the training of non-parametric variables, we view this new paradigm as learning from language.
In practice, scaffolded LMs receive instructions, interface with tools, and receive feedback all in natural language.
Accordingly, we use an LM as an optimizer to interpret this rich language supervision.
In contrast, parametric training has excelled in learning from demonstration (supervised learning), exploration (reinforcement learning), and observation (unsupervised learning), using well-defined loss functions.
Furthermore, it discusses a key capability missing in current scaffolded LMs: the ability to continuously learn across episodes \cite{silver2025era}.
Parametric training inherently suffers from catastrophic forgetting.
Instead, scaffolded LMs are naturally exposed to rich user feedback \cite{openhands, windsurf_editor}.
{\color{changemark}Therefore, we can use an LM optimizer to interpret fuzzy objectives \cite{lee2024aligning} and rich feedback.
}%
Compared to parametric training, it is interpretable, efficient, and compatible with closed-source models.
{\color{changemark}

We discussed \emph{agents that inhabit streams of experience} as AI systems that can continuously learn in settings with significant contextual recurrent.
From a product perspective \cite{cursor2023}, LMs are scaffolded to create product that works with the human in a mixed-autonomy settings.
We hope this survey inspires future research in adaptive AI systems in mixed-autonomy settings, where AI continously adapts to human feedbacks.
}

\section{Limitations}

Our work focuses on organizing a diverse and intricate body of literature, but it does not aim to exhaustively trace how each idea has been applied across all prior work.
For instance, we highlight certain benchmarks that are not consistently used across the papers we discuss \cite{tan2025langprobelanguageprogramsbenchmark}.
This is because these benchmarks are often too challenging for current non-parametric training methods.
This selective approach reflects our intent to extract broader conceptual insights rather than follow any one established framework.
With the recent surge of interest in RL applied to LMs, many earlier methods and assumptions are becoming less central.
As a result, we intentionally de-emphasize some threads of the literature that, while once influential, no longer align with the direction of current research.
This allows us to focus on outlining key future research areas that involve learning from language.

In addition to learning from language, there has been limited work in learning to incorporate visual information, as some concepts—find me this cup—are difficult to convey purely through text.
Multi-modal prompt-based learning has been explored in prior work on robotics \cite{jiang2023vima}, where prompts may include both text and images, e.g., ``bring me \textit{image of the cup}.''
Inspired by this, scaffolded LMs can also interleave textual and visual tokens within their system prompt.
This is particularly beneficial when images convey information more compactly or clearly than text alone.

\Acknowledgements{This work was supported by the National Natural Science Foundation of China (Grant Nos. W2442033, W2442032, and 62461160309).}

\bibliographystyle{scis}
\bibliography{references}

\begin{thebibliography}{100}
\providecommand{\url}[1]{\texttt{#1}}
\providecommand{\urlprefix}{URL }
\providecommand{\bibinfo}[2]{#2}

\bibitem{mixedautonomy}
\bibinfo{author}{Yuchen Cui}, \bibinfo{author}{Siddharth Karamcheti}, \bibinfo{author}{Raj Palleti}, et~al.
\newblock \bibinfo{title}{No, to the right: Online language corrections for robotic manipulation via shared autonomy}.
\newblock In: \bibinfo{booktitle}{Proceedings of the 2023 ACM/IEEE International Conference on Human-Robot Interaction}, \bibinfo{address}{New York, NY, USA}: \bibinfo{publisher}{Association for Computing Machinery}. \bibinfo{year}{2023}, HRI '23.
\newblock \bibinfo{pages}{93--101}.
\newblock \urlprefix\url{https://doi.org/10.1145/3568162.3578623}

\bibitem{silver2025era}
\bibinfo{author}{David Silver}, \bibinfo{author}{Richard~S Sutton}.
\newblock \bibinfo{title}{The era of experience}, \bibinfo{year}{2025}.
\newblock \urlprefix\url{https://storage.googleapis.com/deepmind-media/Era-of-Experience%20/The%20Era%20of%20Experience%20Paper.pdf}

\bibitem{Hadsell2020}
\bibinfo{author}{Raia Hadsell}, \bibinfo{author}{Dushyant Rao}, \bibinfo{author}{Andrei~A Rusu}, et~al.
\newblock \bibinfo{title}{Embracing change: Continual learning in deep neural networks}.
\newblock \bibinfo{journal}{Trends in Cognitive Sciences}, \bibinfo{year}{2020}, \bibinfo{volume}{24}: \bibinfo{pages}{1028--1040}.
\newblock \urlprefix\url{https://www.sciencedirect.com/science/article/pii/S1364661320302199}

\bibitem{openai_charter}
\bibinfo{author}{{OpenAI}}.
\newblock \bibinfo{title}{Openai charter}, \bibinfo{year}{2018}.
\newblock \urlprefix\url{https://openai.com/charter/}, \bibinfo{note}{accessed: 2025-04-25}

\bibitem{openhands}
\bibinfo{author}{Xingyao Wang}, \bibinfo{author}{Boxuan Li}, \bibinfo{author}{Yufan Song}, et~al.
\newblock \bibinfo{title}{{OpenHands: An Open Platform for AI Software Developers as Generalist Agents}}, \bibinfo{year}{2024}.
\newblock \urlprefix\url{https://arxiv.org/abs/2407.16741}

\bibitem{openai2024mlebench}
\bibinfo{author}{Jun~Shern Chan}, \bibinfo{author}{Neil Chowdhury}, \bibinfo{author}{Oliver Jaffe}, et~al.
\newblock \bibinfo{title}{{MLE-bench}: Evaluating machine learning agents on machine learning engineering}.
\newblock \bibinfo{year}{2024}.
\newblock \urlprefix\url{https://arxiv.org/abs/2410.07095}

\bibitem{openai_function_calling}
\bibinfo{author}{{OpenAI}}.
\newblock \bibinfo{title}{Function calling}, \bibinfo{year}{2025}.
\newblock \urlprefix\url{https://platform.openai.com/docs/guides/function-calling}, \bibinfo{note}{accessed: 2025-10-19}

\bibitem{wang2024what}
\bibinfo{author}{Zhiruo Wang}, \bibinfo{author}{Zhoujun Cheng}, \bibinfo{author}{Hao Zhu}, et~al.
\newblock \bibinfo{title}{What are tools anyway? a survey from the language model perspective}.
\newblock In: \bibinfo{booktitle}{First Conference on Language Modeling}, \bibinfo{year}{2024}.
\newblock \urlprefix\url{https://openreview.net/forum?id=Xh1B90iBSR}

\bibitem{textgrad}
\bibinfo{author}{Mert Yuksekgonul}, \bibinfo{author}{Federico Bianchi}, \bibinfo{author}{Joseph Boen}, et~al.
\newblock \bibinfo{title}{Optimizing generative ai by backpropagating language model feedback}.
\newblock \bibinfo{journal}{Nature}, \bibinfo{year}{2025}, \bibinfo{volume}{639}: \bibinfo{pages}{609--616}.
\newblock \urlprefix\url{https://doi.org/10.1038/s41586-025-08661-4}

\bibitem{zhao2024expel}
\bibinfo{author}{Andrew Zhao}, \bibinfo{author}{Daniel Huang}, \bibinfo{author}{Quentin Xu}, et~al.
\newblock \bibinfo{title}{Expel: Llm agents are experiential learners}.
\newblock In: \bibinfo{booktitle}{Proceedings of the AAAI Conference on Artificial Intelligence}, \bibinfo{year}{2024}, volume~\bibinfo{volume}{38}.
\newblock \bibinfo{pages}{19632--19642}

\bibitem{wang2025inducingprogrammaticskillsagentic}
\bibinfo{author}{Zora~Zhiruo Wang}, \bibinfo{author}{Apurva Gandhi}, \bibinfo{author}{Graham Neubig}, et~al.
\newblock \bibinfo{title}{Inducing programmatic skills for agentic tasks}, \bibinfo{year}{2025}.
\newblock \urlprefix\url{https://arxiv.org/abs/2504.06821}

\bibitem{rlhf}
\bibinfo{author}{Long Ouyang}, \bibinfo{author}{Jeffrey Wu}, \bibinfo{author}{Xu~Jiang}, et~al.
\newblock \bibinfo{title}{Training language models to follow instructions with human feedback}.
\newblock In: \bibinfo{booktitle}{Advances in Neural Information Processing Systems 35: Annual Conference on Neural Information Processing Systems 2022, NeurIPS 2022, New Orleans, LA, USA, November 28 - December 9, 2022}, \bibinfo{year}{2022}.
\newblock \urlprefix\url{http://papers.nips.cc/paper\_files/paper/2022/hash/b1efde53be364a73914f58805a001731-Abstract-Conference.html}

\bibitem{reasoning}
\bibinfo{author}{DeepSeek-AI}, \bibinfo{author}{Daya Guo}, \bibinfo{author}{Dejian Yang}, et~al.
\newblock \bibinfo{title}{Deepseek-r1: Incentivizing reasoning capability in llms via reinforcement learning}, \bibinfo{year}{2025}.
\newblock \urlprefix\url{https://arxiv.org/abs/2501.12948}

\bibitem{gpt3}
\bibinfo{author}{Tom Brown}, \bibinfo{author}{Benjamin Mann}, \bibinfo{author}{Nick Ryder}, et~al.
\newblock \bibinfo{title}{Language models are few-shot learners}.
\newblock In: \bibinfo{booktitle}{Advances in Neural Information Processing Systems}, \bibinfo{publisher}{Curran Associates, Inc.}. \bibinfo{year}{2020}, volume~\bibinfo{volume}{33}.
\newblock \bibinfo{pages}{1877--1901}.
\newblock \urlprefix\url{https://proceedings.neurips.cc/paper_files/paper/2020/file/1457c0d6bfcb4967418bfb8ac142f64a-Paper.pdf}

\bibitem{lecun2015deep}
\bibinfo{author}{Yann LeCun}, \bibinfo{author}{Yoshua Bengio}, \bibinfo{author}{Geoffrey Hinton}.
\newblock \bibinfo{title}{Deep learning}.
\newblock \bibinfo{journal}{nature}, \bibinfo{year}{2015}, \bibinfo{volume}{521}: \bibinfo{pages}{436}

\bibitem{llfbench}
\bibinfo{author}{Ching-An Cheng}, \bibinfo{author}{Andrey Kolobov}, \bibinfo{author}{Dipendra Misra}, et~al.
\newblock \bibinfo{title}{Llf-bench: Benchmark for interactive learning from language feedback}, \bibinfo{year}{2023}.
\newblock \urlprefix\url{https://arxiv.org/abs/2312.06853}

\bibitem{claude}
\bibinfo{author}{{Anthropic}}.
\newblock \bibinfo{title}{Claude}, \bibinfo{year}{2024}.
\newblock \urlprefix\url{https://www.anthropic.com/claude}, \bibinfo{note}{accessed: Month, Day, 2024}

\bibitem{chatgpt}
\bibinfo{author}{{OpenAI}}.
\newblock \bibinfo{title}{Chatgpt}, \bibinfo{year}{2024}.
\newblock \urlprefix\url{https://openai.com/chatgpt}, \bibinfo{note}{accessed: Month, Day, 2024}

\bibitem{compound-ai-blog}
\bibinfo{author}{Matei Zaharia}, \bibinfo{author}{Omar Khattab}, \bibinfo{author}{Lingjiao Chen}, et~al.
\newblock \bibinfo{title}{The shift from models to compound ai systems}, \bibinfo{year}{2024}.
\newblock \urlprefix\url{https://bair.berkeley.edu/blog/2024/02/18/compound-ai-systems/}

\bibitem{dspy}
\bibinfo{author}{Omar Khattab}, \bibinfo{author}{Arnav Singhvi}, \bibinfo{author}{Paridhi Maheshwari}, et~al.
\newblock \bibinfo{title}{Dspy: Compiling declarative language model calls into self-improving pipelines}.
\newblock \bibinfo{journal}{CoRR}, \bibinfo{year}{2023}, \bibinfo{volume}{abs/2310.03714}.
\newblock \urlprefix\url{https://doi.org/10.48550/arXiv.2310.03714}

\bibitem{ape}
\bibinfo{author}{Yongchao Zhou}, \bibinfo{author}{Andrei~Ioan Muresanu}, \bibinfo{author}{Ziwen Han}, et~al.
\newblock \bibinfo{title}{Large language models are human-level prompt engineers}.
\newblock In: \bibinfo{booktitle}{The Eleventh International Conference on Learning Representations, {ICLR} 2023, Kigali, Rwanda, May 1-5, 2023}, \bibinfo{publisher}{OpenReview.net}. \bibinfo{year}{2023}.
\newblock \urlprefix\url{https://openreview.net/forum?id=92gvk82DE-}

\bibitem{zhao2025compromisedprompts}
\bibinfo{author}{Andrew Zhao}, \bibinfo{author}{Reshmi Ghosh}, \bibinfo{author}{Vitor Carvalho}, et~al.
\newblock \bibinfo{title}{Are my optimized prompts compromised? exploring vulnerabilities of llm-based optimizers}, \bibinfo{year}{2025}.
\newblock \urlprefix\url{https://arxiv.org/abs/2510.14381}

\bibitem{openai2025agents}
\bibinfo{author}{{OpenAI}}.
\newblock \bibinfo{title}{New tools for building agents}, \bibinfo{year}{2025}.
\newblock \urlprefix\url{https://openai.com/index/new-tools-for-building-agents/}, \bibinfo{note}{accessed: 2025-04-17}

\bibitem{yaoshunyuthesis}
\bibinfo{author}{Shunyu Yao}.
\newblock \bibinfo{title}{Language agents: From next-token prediction to digital automation}.
\newblock Ph.D. thesis, \bibinfo{school}{Princeton University}, \bibinfo{address}{Princeton, NJ, USA}, \bibinfo{year}{2024}.
\newblock \urlprefix\url{https://www.cs.princeton.edu/people/profile/yaos}, \bibinfo{note}{doctor of Philosophy (Ph.D.) Thesis}

\bibitem{rise-and-potential}
\bibinfo{author}{Zhiheng Xi}, \bibinfo{author}{Wenxiang Chen}, \bibinfo{author}{Xin Guo}, et~al.
\newblock \bibinfo{title}{The rise and potential of large language model based agents: a survey}.
\newblock \bibinfo{journal}{Science China Information Sciences}, \bibinfo{year}{2025}, \bibinfo{volume}{68}: \bibinfo{pages}{121101}.
\newblock \urlprefix\url{https://link.springer.com/article/10.1007/s11432-024-4222-0}

\bibitem{landscape-emerging-agent}
\bibinfo{author}{Tula Masterman}, \bibinfo{author}{Sandi Besen}, \bibinfo{author}{Mason Sawtell}, et~al.
\newblock \bibinfo{title}{The landscape of emerging {AI} agent architectures for reasoning, planning, and tool calling: {A} survey}.
\newblock \bibinfo{journal}{CoRR}, \bibinfo{year}{2024}, \bibinfo{volume}{abs/2404.11584}.
\newblock \urlprefix\url{https://doi.org/10.48550/arXiv.2404.11584}

\bibitem{trace}
\bibinfo{author}{Ching{-}An Cheng}, \bibinfo{author}{Allen Nie}, \bibinfo{author}{Adith Swaminathan}.
\newblock \bibinfo{title}{Trace is the new autodiff - unlocking efficient optimization of computational workflows}.
\newblock \bibinfo{journal}{CoRR}, \bibinfo{year}{2024}, \bibinfo{volume}{abs/2406.16218}.
\newblock \urlprefix\url{https://doi.org/10.48550/arXiv.2406.16218}

\bibitem{adalflow}
\bibinfo{author}{Li~Yin}.
\newblock \bibinfo{title}{{AdalFlow: The Library for Large Language Model (LLM) Applications}}, \bibinfo{year}{2024}.
\newblock \urlprefix\url{https://github.com/SylphAI-Inc/AdalFlow}

\bibitem{lin2024llmbasedoptimizationcompoundai}
\bibinfo{author}{Matthieu Lin}, \bibinfo{author}{Jenny Sheng}, \bibinfo{author}{Andrew Zhao}, et~al.
\newblock \bibinfo{title}{Llm-based optimization of compound ai systems: A survey}, \bibinfo{year}{2024}.
\newblock \urlprefix\url{https://arxiv.org/abs/2410.16392}

\bibitem{yao2025second}
\bibinfo{author}{Shunyu Yao}.
\newblock \bibinfo{title}{The second half}, \bibinfo{year}{2025}.
\newblock \urlprefix\url{https://ysymyth.github.io/The-Second-Half/}, \bibinfo{note}{blog post}

\bibitem{llm-autonomous-agents-survey}
\bibinfo{author}{Lei Wang}, \bibinfo{author}{Chen Ma}, \bibinfo{author}{Xueyang Feng}, et~al.
\newblock \bibinfo{title}{A survey on large language model based autonomous agents}.
\newblock \bibinfo{journal}{Frontiers of Computer Science}, \bibinfo{year}{2024}, \bibinfo{volume}{18}: \bibinfo{pages}{186345}.
\newblock \urlprefix\url{https://link.springer.com/article/10.1007/s11704-024-40231-1}

\bibitem{feedback-mechanism-llm-agents}
\bibinfo{author}{Zhipeng Liu}, \bibinfo{author}{Xuefeng Bai}, \bibinfo{author}{Kehai Chen}, et~al.
\newblock \bibinfo{title}{A survey on the feedback mechanism of {LLM}-based {AI} agents}.
\newblock In: \bibinfo{booktitle}{Proceedings of the Thirty-Fourth International Joint Conference on Artificial Intelligence (IJCAI)}, \bibinfo{year}{2025}.
\newblock \bibinfo{pages}{10582--10592}

\bibitem{deepseekai2025deepseekr1incentivizingreasoningcapability}
\bibinfo{author}{DeepSeek-AI}.
\newblock \bibinfo{title}{Deepseek-r1: Incentivizing reasoning capability in llms via reinforcement learning}, \bibinfo{year}{2025}.
\newblock \urlprefix\url{https://arxiv.org/abs/2501.12948}

\bibitem{tie2025posttraining}
\bibinfo{author}{Guiyao Tie}, \bibinfo{author}{Zeli Zhao}, \bibinfo{author}{Dingjie Song}, et~al.
\newblock \bibinfo{title}{A survey on post-training of large language models}, \bibinfo{year}{2025}.
\newblock \urlprefix\url{https://arxiv.org/abs/2503.06072}

\bibitem{lai-etal-2025-survey}
\bibinfo{author}{Kenny~Shijian Lai}, \bibinfo{author}{Jasur Mirzakhalov}, \bibinfo{author}{Karan Singla}, et~al.
\newblock \bibinfo{title}{A survey of post-training scaling in large language models}.
\newblock In: \bibinfo{booktitle}{Proceedings of the 63rd Annual Meeting of the Association for Computational Linguistics (Volume 1: Long Papers)}, \bibinfo{address}{Vienna, Austria}: \bibinfo{publisher}{Association for Computational Linguistics}. \bibinfo{year}{2025}.
\newblock \bibinfo{pages}{2478--2510}.
\newblock \urlprefix\url{https://aclanthology.org/2025.acl-long.140}

\bibitem{du2025surveyoptimizationlargelanguage}
\bibinfo{author}{Shangheng Du}, \bibinfo{author}{Jiabao Zhao}, \bibinfo{author}{Jinxin Shi}, et~al.
\newblock \bibinfo{title}{A survey on the optimization of large language model-based agents}, \bibinfo{year}{2025}.
\newblock \urlprefix\url{https://arxiv.org/abs/2503.12434}

\bibitem{cobbe2021gsm8k}
\bibinfo{author}{Karl Cobbe}, \bibinfo{author}{Vineet Kosaraju}, \bibinfo{author}{Mohammad Bavarian}, et~al.
\newblock \bibinfo{title}{Training verifiers to solve math word problems}.
\newblock \bibinfo{journal}{arXiv preprint arXiv:2110.14168}, \bibinfo{year}{2021}

\bibitem{promptagent}
\bibinfo{author}{Xinyuan Wang}, \bibinfo{author}{Chenxi Li}, \bibinfo{author}{Zhen Wang}, et~al.
\newblock \bibinfo{title}{Promptagent: Strategic planning with language models enables expert-level prompt optimization}.
\newblock In: \bibinfo{booktitle}{The Twelfth International Conference on Learning Representations, {ICLR} 2024, Vienna, Austria, May 7-11, 2024}, \bibinfo{publisher}{OpenReview.net}. \bibinfo{year}{2024}.
\newblock \urlprefix\url{https://openreview.net/forum?id=22pyNMuIoa}

\bibitem{protegi}
\bibinfo{author}{Reid Pryzant}, \bibinfo{author}{Dan Iter}, \bibinfo{author}{Jerry Li}, et~al.
\newblock \bibinfo{title}{Automatic prompt optimization with "gradient descent" and beam search}.
\newblock In: \bibinfo{booktitle}{Proceedings of the 2023 Conference on Empirical Methods in Natural Language Processing, {EMNLP} 2023, Singapore, December 6-10, 2023}, \bibinfo{publisher}{Association for Computational Linguistics}. \bibinfo{year}{2023}.
\newblock \bibinfo{pages}{7957--7968}.
\newblock \urlprefix\url{https://doi.org/10.18653/v1/2023.emnlp-main.494}

\bibitem{zhao2024diverctdiversityenhancedredteaming}
\bibinfo{author}{Andrew Zhao}, \bibinfo{author}{Quentin Xu}, \bibinfo{author}{Matthieu Lin}, et~al.
\newblock \bibinfo{title}{Diver-ct: Diversity-enhanced red teaming large language model assistants with relaxing constraints}, \bibinfo{year}{2024}.
\newblock \urlprefix\url{https://arxiv.org/abs/2405.19026}

\bibitem{deng2022rlpromptoptimizingdiscretetext}
\bibinfo{author}{Mingkai Deng}, \bibinfo{author}{Jianyu Wang}, \bibinfo{author}{Cheng-Ping Hsieh}, et~al.
\newblock \bibinfo{title}{Rlprompt: Optimizing discrete text prompts with reinforcement learning}, \bibinfo{year}{2022}.
\newblock \urlprefix\url{https://arxiv.org/abs/2205.12548}

\bibitem{prewrite}
\bibinfo{author}{Weize Kong}, \bibinfo{author}{Spurthi~Amba Hombaiah}, \bibinfo{author}{Mingyang Zhang}, et~al.
\newblock \bibinfo{title}{Prewrite: Prompt rewriting with reinforcement learning}, \bibinfo{year}{2024}.
\newblock \urlprefix\url{https://arxiv.org/abs/2401.08189}

\bibitem{stableprompt}
\bibinfo{author}{Minchan Kwon}, \bibinfo{author}{Gaeun Kim}, \bibinfo{author}{Jongsuk Kim}, et~al.
\newblock \bibinfo{title}{Stableprompt: Automatic prompt tuning using reinforcement learning for large language models}, \bibinfo{year}{2024}.
\newblock \urlprefix\url{https://arxiv.org/abs/2410.07652}

\bibitem{tempera}
\bibinfo{author}{Tianjun Zhang}, \bibinfo{author}{Xuezhi Wang}, \bibinfo{author}{Denny Zhou}, et~al.
\newblock \bibinfo{title}{{TEMPERA:} test-time prompting via reinforcement learning}.
\newblock \bibinfo{journal}{CoRR}, \bibinfo{year}{2022}, \bibinfo{volume}{abs/2211.11890}.
\newblock \urlprefix\url{https://doi.org/10.48550/arXiv.2211.11890}

\bibitem{zhao2025absolutezeroreinforcedselfplay}
\bibinfo{author}{Andrew Zhao}, \bibinfo{author}{Yiran Wu}, \bibinfo{author}{Yang Yue}, et~al.
\newblock \bibinfo{title}{Absolute zero: Reinforced self-play reasoning with zero data}, \bibinfo{year}{2025}.
\newblock \urlprefix\url{https://arxiv.org/abs/2505.03335}

\bibitem{wang2025reinforcementlearningreasoninglarge}
\bibinfo{author}{Yiping Wang}, \bibinfo{author}{Qing Yang}, \bibinfo{author}{Zhiyuan Zeng}, et~al.
\newblock \bibinfo{title}{Reinforcement learning for reasoning in large language models with one training example}, \bibinfo{year}{2025}.
\newblock \urlprefix\url{https://arxiv.org/abs/2504.20571}

\bibitem{wei2025swerl}
\bibinfo{author}{Yuxiang Wei}, \bibinfo{author}{Olivier Duchenne}, \bibinfo{author}{Jade Copet}, et~al.
\newblock \bibinfo{title}{Swe-rl: Advancing llm reasoning via reinforcement learning on open software evolution}.
\newblock \bibinfo{journal}{arXiv preprint arXiv:2502.18449}, \bibinfo{year}{2025}

\bibitem{liu2023is}
\bibinfo{author}{Jiawei Liu}, \bibinfo{author}{Chunqiu~Steven Xia}, \bibinfo{author}{Yuyao Wang}, et~al.
\newblock \bibinfo{title}{Is your code generated by chat{GPT} really correct? rigorous evaluation of large language models for code generation}.
\newblock In: \bibinfo{booktitle}{Thirty-seventh Conference on Neural Information Processing Systems}, \bibinfo{year}{2023}.
\newblock \urlprefix\url{https://openreview.net/forum?id=1qvx610Cu7}

\bibitem{agentless}
\bibinfo{author}{Chunqiu~Steven Xia}, \bibinfo{author}{Yinlin Deng}, \bibinfo{author}{Soren Dunn}, et~al.
\newblock \bibinfo{title}{Agentless: Demystifying llm-based software engineering agents}.
\newblock \bibinfo{journal}{CoRR}, \bibinfo{year}{2024}, \bibinfo{volume}{abs/2407.01489}.
\newblock \urlprefix\url{https://doi.org/10.48550/arXiv.2407.01489}

\bibitem{jimenez2024swebench}
\bibinfo{author}{Carlos~E Jimenez}, \bibinfo{author}{John Yang}, \bibinfo{author}{Alexander Wettig}, et~al.
\newblock \bibinfo{title}{{SWE}-bench: Can language models resolve real-world github issues?}
\newblock In: \bibinfo{booktitle}{The Twelfth International Conference on Learning Representations}, \bibinfo{year}{2024}.
\newblock \urlprefix\url{https://openreview.net/forum?id=VTF8yNQM66}

\bibitem{zhuo2025bigcodebench}
\bibinfo{author}{Terry~Yue Zhuo}, \bibinfo{author}{Vu~Minh Chien}, \bibinfo{author}{Jenny Chim}, et~al.
\newblock \bibinfo{title}{Bigcodebench: Benchmarking code generation with diverse function calls and complex instructions}.
\newblock In: \bibinfo{booktitle}{The Thirteenth International Conference on Learning Representations}, \bibinfo{year}{2025}.
\newblock \urlprefix\url{https://openreview.net/forum?id=YrycTjllL0}

\bibitem{codereasoning}
\bibinfo{author}{Alex Gu}, \bibinfo{author}{Baptiste Roziere}, \bibinfo{author}{Hugh~James Leather}, et~al.
\newblock \bibinfo{title}{{CRUXE}val: A benchmark for code reasoning, understanding and execution}.
\newblock In: \bibinfo{booktitle}{Proceedings of the 41st International Conference on Machine Learning}, \bibinfo{publisher}{PMLR}. \bibinfo{year}{2024}, \emph{\bibinfo{series}{Proceedings of Machine Learning Research}}, volume \bibinfo{volume}{235}.
\newblock \bibinfo{pages}{16568--16621}.
\newblock \urlprefix\url{https://proceedings.mlr.press/v235/gu24c.html}

\bibitem{mathhendrycks}
\bibinfo{author}{Dan Hendrycks}, \bibinfo{author}{Collin Burns}, \bibinfo{author}{Saurav Kadavath}, et~al.
\newblock \bibinfo{title}{Measuring mathematical problem solving with the {MATH} dataset}.
\newblock In: \bibinfo{booktitle}{Thirty-fifth Conference on Neural Information Processing Systems Datasets and Benchmarks Track (Round 2)}, \bibinfo{year}{2021}.
\newblock \urlprefix\url{https://openreview.net/forum?id=7Bywt2mQsCe}

\bibitem{hendrycks2021measuring}
\bibinfo{author}{Dan Hendrycks}, \bibinfo{author}{Collin Burns}, \bibinfo{author}{Steven Basart}, et~al.
\newblock \bibinfo{title}{Measuring massive multitask language understanding}.
\newblock In: \bibinfo{booktitle}{International Conference on Learning Representations}, \bibinfo{year}{2021}.
\newblock \urlprefix\url{https://openreview.net/forum?id=d7KBjmI3GmQ}

\bibitem{wei2025verifiersrule}
\bibinfo{author}{Jason Wei}.
\newblock \bibinfo{title}{Asymmetry of verification and verifier's rule}, \bibinfo{year}{2025}.
\newblock \urlprefix\url{https://www.jasonwei.net/blog/asymmetry-of-verification-and-verifiers-law}, \bibinfo{note}{accessed: 2025-10-20}

\bibitem{welleck2024from}
\bibinfo{author}{Sean Welleck}, \bibinfo{author}{Amanda Bertsch}, \bibinfo{author}{Matthew Finlayson}, et~al.
\newblock \bibinfo{title}{From decoding to meta-generation: Inference-time algorithms for large language models}.
\newblock \bibinfo{journal}{Transactions on Machine Learning Research}, \bibinfo{year}{2024}.
\newblock \urlprefix\url{https://openreview.net/forum?id=eskQMcIbMS}, \bibinfo{note}{survey Certification}

\bibitem{adas}
\bibinfo{author}{Shengran Hu}, \bibinfo{author}{Cong Lu}, \bibinfo{author}{Jeff Clune}.
\newblock \bibinfo{title}{Automated design of agentic systems}.
\newblock \bibinfo{journal}{CoRR}, \bibinfo{year}{2024}, \bibinfo{volume}{abs/2408.08435}.
\newblock \urlprefix\url{https://doi.org/10.48550/arXiv.2408.08435}

\bibitem{zhang2025aflow}
\bibinfo{author}{Jiayi Zhang}, \bibinfo{author}{Jinyu Xiang}, \bibinfo{author}{Zhaoyang Yu}, et~al.
\newblock \bibinfo{title}{{AF}low: Automating agentic workflow generation}.
\newblock In: \bibinfo{booktitle}{The Thirteenth International Conference on Learning Representations}, \bibinfo{year}{2025}.
\newblock \urlprefix\url{https://openreview.net/forum?id=z5uVAKwmjf}

\bibitem{archon}
\bibinfo{author}{Jon Saad-Falcon}, \bibinfo{author}{Adrian~Gamarra Lafuente}, \bibinfo{author}{Shlok Natarajan}, et~al.
\newblock \bibinfo{title}{Archon: An architecture search framework for inference-time techniques}, \bibinfo{year}{2024}.
\newblock \urlprefix\url{https://arxiv.org/abs/2409.15254}

\bibitem{silver2016mastering}
\bibinfo{author}{David Silver}, \bibinfo{author}{Aja Huang}, \bibinfo{author}{Chris~J Maddison}, et~al.
\newblock \bibinfo{title}{Mastering the game of go with deep neural networks and tree search}.
\newblock \bibinfo{journal}{Nature}, \bibinfo{year}{2016}, \bibinfo{volume}{529}: \bibinfo{pages}{484--489}

\bibitem{grattafiori2024llama3herdmodels}
\bibinfo{author}{Aaron Grattafiori}, \bibinfo{author}{Abhimanyu Dubey}, \bibinfo{author}{Abhinav Jauhri}, et~al.
\newblock \bibinfo{title}{The llama 3 herd of models}, \bibinfo{year}{2024}.
\newblock \urlprefix\url{https://arxiv.org/abs/2407.21783}

\bibitem{kundel2025harmony}
\bibinfo{author}{Dominik Kundel}.
\newblock \bibinfo{title}{Openai harmony response format}.
\newblock \bibinfo{howpublished}{\url{https://cookbook.openai.com/articles/openai-harmony}}, \bibinfo{year}{2025}.
\newblock \bibinfo{note}{OpenAI Cookbook; accessed 2025-10-21}

\bibitem{wallace2024instructionhierarchytrainingllms}
\bibinfo{author}{Eric Wallace}, \bibinfo{author}{Kai Xiao}, \bibinfo{author}{Reimar Leike}, et~al.
\newblock \bibinfo{title}{The instruction hierarchy: Training llms to prioritize privileged instructions}, \bibinfo{year}{2024}.
\newblock \urlprefix\url{https://arxiv.org/abs/2404.13208}

\bibitem{openai_codex_cloud_internet_access}
\bibinfo{author}{{OpenAI}}.
\newblock \bibinfo{title}{Codex cloud: Internet access}, \bibinfo{year}{2025}.
\newblock \urlprefix\url{https://developers.openai.com/codex/cloud/internet-access}, \bibinfo{note}{accessed: October 22, 2025}

\bibitem{chao2023jailbreaking}
\bibinfo{author}{Patrick Chao}, \bibinfo{author}{Alexander Robey}, \bibinfo{author}{Edgar Dobriban}, et~al.
\newblock \bibinfo{title}{Jailbreaking black box large language models in twenty queries}, \bibinfo{year}{2023}.
\newblock \urlprefix\url{https://arxiv.org/abs/2310.08419}

\bibitem{breunig2025claude}
\bibinfo{author}{Drew Breunig}.
\newblock \bibinfo{title}{Claude's system prompt: Chatbots are more than just models}, \bibinfo{year}{2025}.
\newblock \urlprefix\url{https://www.dbreunig.com/2025/05/07/claude-s-system-prompt-chatbots-are-more-than-just-models.html}, \bibinfo{note}{accessed: 2025-05-11}

\bibitem{li2025tamingknowledgeconflictslanguage}
\bibinfo{author}{Gaotang Li}, \bibinfo{author}{Yuzhong Chen}, \bibinfo{author}{Hanghang Tong}.
\newblock \bibinfo{title}{Taming knowledge conflicts in language models}, \bibinfo{year}{2025}.
\newblock \urlprefix\url{https://arxiv.org/abs/2503.10996}

\bibitem{knowledge-conflict-survey}
\bibinfo{author}{Rongwu Xu}, \bibinfo{author}{Zehan Qi}, \bibinfo{author}{Zhijiang Guo}, et~al.
\newblock \bibinfo{title}{Knowledge conflicts for llms: {A} survey}.
\newblock In: \bibinfo{booktitle}{Proceedings of the 2024 Conference on Empirical Methods in Natural Language Processing, {EMNLP} 2024, Miami, FL, USA, November 12-16, 2024}, \bibinfo{publisher}{Association for Computational Linguistics}. \bibinfo{year}{2024}.
\newblock \bibinfo{pages}{8541--8565}.
\newblock \urlprefix\url{https://aclanthology.org/2024.emnlp-main.486}

\bibitem{asgeirtj2025claude}
\bibinfo{author}{Ásgeir Thor~Johnson}.
\newblock \bibinfo{title}{Claude system prompt leak}, \bibinfo{year}{2025}.
\newblock \urlprefix\url{https://github.com/asgeirtj/system_prompts_leaks/blob/main/claude.txt}, \bibinfo{note}{accessed: 2025-05-11}

\bibitem{anthropic2025thinktool}
\bibinfo{author}{{Anthropic}}.
\newblock \bibinfo{title}{The "think" tool: Enabling claude to stop and think in complex tool use situations}, \bibinfo{year}{2025}.
\newblock \urlprefix\url{https://www.anthropic.com/engineering/claude-think-tool}, \bibinfo{note}{accessed: 2025-04-22}

\bibitem{langchain_human_tool}
\bibinfo{author}{LangChain}.
\newblock \bibinfo{title}{Human as a tool}, \bibinfo{year}{2025}.
\newblock \urlprefix\url{https://python.langchain.com/docs/integrations/tools/human_tools/}, \bibinfo{note}{accessed: 2025-05-11}

\bibitem{openai_agents_sdk}
\bibinfo{author}{{OpenAI}}.
\newblock \bibinfo{title}{Openai agents sdk}, \bibinfo{year}{2025}.
\newblock \urlprefix\url{https://github.com/openai/openai-agents-python}, \bibinfo{note}{accessed: 2025-05-11}

\bibitem{mccoy2023embersautoregressionunderstandinglarge}
\bibinfo{author}{R~Thomas McCoy}, \bibinfo{author}{Shunyu Yao}, \bibinfo{author}{Dan Friedman}, et~al.
\newblock \bibinfo{title}{Embers of autoregression: Understanding large language models through the problem they are trained to solve}, \bibinfo{year}{2023}.
\newblock \urlprefix\url{https://arxiv.org/abs/2309.13638}

\bibitem{grosse2023studyinglargelanguagemodel}
\bibinfo{author}{Roger Grosse}, \bibinfo{author}{Juhan Bae}, \bibinfo{author}{Cem Anil}, et~al.
\newblock \bibinfo{title}{Studying large language model generalization with influence functions}, \bibinfo{year}{2023}.
\newblock \urlprefix\url{https://arxiv.org/abs/2308.03296}, \bibinfo{note}{see page 45 for relevant details}

\bibitem{openai2025gptoss120bgptoss20bmodel}
\bibinfo{author}{OpenAI}.
\newblock \bibinfo{title}{gpt-oss-120b \& gpt-oss-20b model card}, \bibinfo{year}{2025}.
\newblock \urlprefix\url{https://arxiv.org/abs/2508.10925}

\bibitem{nakano2022webgptbrowserassistedquestionansweringhuman}
\bibinfo{author}{Reiichiro Nakano}, \bibinfo{author}{Jacob Hilton}, \bibinfo{author}{Suchir Balaji}, et~al.
\newblock \bibinfo{title}{Webgpt: Browser-assisted question-answering with human feedback}, \bibinfo{year}{2022}.
\newblock \urlprefix\url{https://arxiv.org/abs/2112.09332}

\bibitem{rag}
\bibinfo{author}{Patrick Lewis}, \bibinfo{author}{Ethan Perez}, \bibinfo{author}{Aleksandra Piktus}, et~al.
\newblock \bibinfo{title}{Retrieval-augmented generation for knowledge-intensive nlp tasks}.
\newblock In: \bibinfo{booktitle}{Proceedings of the 34th International Conference on Neural Information Processing Systems}, \bibinfo{address}{Red Hook, NY, USA}: \bibinfo{publisher}{Curran Associates Inc.}. \bibinfo{year}{2020}, NIPS '20

\bibitem{basic}
\bibinfo{author}{Gautier Izacard}, \bibinfo{author}{Edouard Grave}.
\newblock \bibinfo{title}{Leveraging passage retrieval with generative models for open domain question answering}.
\newblock In: \bibinfo{booktitle}{Proceedings of the 16th Conference of the European Chapter of the Association for Computational Linguistics: Main Volume, {EACL} 2021, Online, April 19 - 23, 2021}, \bibinfo{publisher}{Association for Computational Linguistics}. \bibinfo{year}{2021}.
\newblock \bibinfo{pages}{874--880}.
\newblock \urlprefix\url{https://doi.org/10.18653/v1/2021.eacl-main.74}

\bibitem{ralm}
\bibinfo{author}{Ori Ram}, \bibinfo{author}{Yoav Levine}, \bibinfo{author}{Itay Dalmedigos}, et~al.
\newblock \bibinfo{title}{In-context retrieval-augmented language models}.
\newblock \bibinfo{journal}{Trans. Assoc. Comput. Linguistics}, \bibinfo{year}{2023}, \bibinfo{volume}{11}: \bibinfo{pages}{1316--1331}.
\newblock \urlprefix\url{https://doi.org/10.1162/tacl\_a\_00605}

\bibitem{context-retrieval}
\bibinfo{author}{Anthropic}.
\newblock \bibinfo{title}{Introducing contextual retrieval}, \bibinfo{year}{2024}.
\newblock \urlprefix\url{https://www.anthropic.com/news/contextual-retrieval}, \bibinfo{note}{accessed: 2024-10-01}

\bibitem{hover}
\bibinfo{author}{Yichen Jiang}, \bibinfo{author}{Shikha Bordia}, \bibinfo{author}{Zheng Zhong}, et~al.
\newblock \bibinfo{title}{Hover: {A} dataset for many-hop fact extraction and claim verification}.
\newblock In: \bibinfo{booktitle}{Findings of the Association for Computational Linguistics: {EMNLP} 2020, Online Event, 16-20 November 2020}, \bibinfo{publisher}{Association for Computational Linguistics}. \bibinfo{year}{2020}, \emph{\bibinfo{series}{Findings of {ACL}}}, volume \bibinfo{volume}{{EMNLP} 2020}.
\newblock \bibinfo{pages}{3441--3460}.
\newblock \urlprefix\url{https://doi.org/10.18653/v1/2020.findings-emnlp.309}

\bibitem{generative-agents}
\bibinfo{author}{Joon~Sung Park}, \bibinfo{author}{Joseph~C O'Brien}, \bibinfo{author}{Carrie~Jun Cai}, et~al.
\newblock \bibinfo{title}{Generative agents: Interactive simulacra of human behavior}.
\newblock In: \bibinfo{booktitle}{Proceedings of the 36th Annual {ACM} Symposium on User Interface Software and Technology, {UIST} 2023, San Francisco, CA, USA, 29 October 2023- 1 November 2023}, \bibinfo{publisher}{{ACM}}. \bibinfo{year}{2023}.
\newblock \bibinfo{pages}{2:1--2:22}.
\newblock \urlprefix\url{https://doi.org/10.1145/3586183.3606763}

\bibitem{openai2025operator}
\bibinfo{author}{{OpenAI}}.
\newblock \bibinfo{title}{Operator system card}.
\newblock \bibinfo{type}{Technical Report}, \bibinfo{institution}{OpenAI}, \bibinfo{year}{2025}.
\newblock \urlprefix\url{https://cdn.openai.com/operator_system_card.pdf}, \bibinfo{note}{accessed: 2025-04-28}

\bibitem{codeact}
\bibinfo{author}{Xingyao Wang}, \bibinfo{author}{Yangyi Chen}, \bibinfo{author}{Lifan Yuan}, et~al.
\newblock \bibinfo{title}{Executable code actions elicit better {LLM} agents}.
\newblock In: \bibinfo{booktitle}{Forty-first International Conference on Machine Learning, {ICML} 2024, Vienna, Austria, July 21-27, 2024}, \bibinfo{publisher}{OpenReview.net}. \bibinfo{year}{2024}.
\newblock \urlprefix\url{https://openreview.net/forum?id=jJ9BoXAfFa}

\bibitem{ma2025pouproofuse}
\bibinfo{author}{Shengjie Ma}, \bibinfo{author}{Chenlong Deng}, \bibinfo{author}{Jiaxin Mao}, et~al.
\newblock \bibinfo{title}{Pou: Proof-of-use to counter tool-call hacking in deepresearch agents}, \bibinfo{year}{2025}.
\newblock \urlprefix\url{https://arxiv.org/abs/2510.10931}

\bibitem{langchain2025agentframeworks}
\bibinfo{author}{LangChain}.
\newblock \bibinfo{title}{How to think about agent frameworks}, \bibinfo{year}{2025}.
\newblock \urlprefix\url{https://blog.langchain.dev/how-to-think-about-agent-frameworks/}, \bibinfo{note}{accessed: 2025-04-22}

\bibitem{lu2024aiscientist}
\bibinfo{author}{Chris Lu}, \bibinfo{author}{Cong Lu}, \bibinfo{author}{Robert~Tjarko Lange}, et~al.
\newblock \bibinfo{title}{The {AI} {S}cientist: Towards fully automated open-ended scientific discovery}.
\newblock \bibinfo{journal}{arXiv preprint arXiv:2408.06292}, \bibinfo{year}{2024}

\bibitem{aiscientist_v2}
\bibinfo{author}{Yutaro Yamada}, \bibinfo{author}{Robert~Tjarko Lange}, \bibinfo{author}{Cong Lu}, et~al.
\newblock \bibinfo{title}{The ai scientist-v2: Workshop-level automated scientific discovery via agentic tree search}.
\newblock \bibinfo{journal}{arXiv preprint arXiv:2504.08066}, \bibinfo{year}{2025}

\bibitem{pan2024trainingsoftwareengineeringagents}
\bibinfo{author}{Jiayi Pan}, \bibinfo{author}{Xingyao Wang}, \bibinfo{author}{Graham Neubig}, et~al.
\newblock \bibinfo{title}{Training software engineering agents and verifiers with swe-gym}, \bibinfo{year}{2024}.
\newblock \urlprefix\url{https://arxiv.org/abs/2412.21139}

\bibitem{snell2024scalingllmtesttimecompute}
\bibinfo{author}{Charlie Snell}, \bibinfo{author}{Jaehoon Lee}, \bibinfo{author}{Kelvin Xu}, et~al.
\newblock \bibinfo{title}{Scaling llm test-time compute optimally can be more effective than scaling model parameters}, \bibinfo{year}{2024}.
\newblock \urlprefix\url{https://arxiv.org/abs/2408.03314}

\bibitem{chen2024llmcallsneedscaling}
\bibinfo{author}{Lingjiao Chen}, \bibinfo{author}{Jared~Quincy Davis}, \bibinfo{author}{Boris Hanin}, et~al.
\newblock \bibinfo{title}{Are more llm calls all you need? towards scaling laws of compound inference systems}, \bibinfo{year}{2024}.
\newblock \urlprefix\url{https://arxiv.org/abs/2403.02419}

\bibitem{self-consistency}
\bibinfo{author}{Xuezhi Wang}, \bibinfo{author}{Jason Wei}, \bibinfo{author}{Dale Schuurmans}, et~al.
\newblock \bibinfo{title}{Self-consistency improves chain of thought reasoning in language models}.
\newblock In: \bibinfo{booktitle}{The Eleventh International Conference on Learning Representations, {ICLR} 2023, Kigali, Rwanda, May 1-5, 2023}, \bibinfo{publisher}{OpenReview.net}. \bibinfo{year}{2023}.
\newblock \urlprefix\url{https://openreview.net/forum?id=1PL1NIMMrw}

\bibitem{tot}
\bibinfo{author}{Shunyu Yao}, \bibinfo{author}{Dian Yu}, \bibinfo{author}{Jeffrey Zhao}, et~al.
\newblock \bibinfo{title}{Tree of thoughts: Deliberate problem solving with large language models}.
\newblock In: \bibinfo{booktitle}{Advances in Neural Information Processing Systems 36: Annual Conference on Neural Information Processing Systems 2023, NeurIPS 2023, New Orleans, LA, USA, December 10 - 16, 2023}, \bibinfo{year}{2023}.
\newblock \urlprefix\url{http://papers.nips.cc/paper\_files/paper/2023/hash/271db9922b8d1f4dd7aaef84ed5ac703-Abstract-Conference.html}

\bibitem{windsurf_editor}
\bibinfo{author}{{Windsurf}}.
\newblock \bibinfo{title}{Windsurf editor}.
\newblock \urlprefix\url{https://windsurf.com/editor}, \bibinfo{note}{accessed: 2025-05-10}

\bibitem{cursor2023}
\bibinfo{author}{{Cursor}}.
\newblock \bibinfo{title}{Cursor: The ai-powered code editor}, \bibinfo{year}{2023}.
\newblock \urlprefix\url{https://www.cursor.com/}, \bibinfo{note}{accessed: 2025-05-14}

\bibitem{microsoft2025copilot}
\bibinfo{author}{Microsoft}.
\newblock \bibinfo{title}{Copilot (gpt-4) [large language model]}, \bibinfo{year}{2025}.
\newblock \urlprefix\url{https://copilot.microsoft.com/}, \bibinfo{note}{accessed: 2025-05-13}

\bibitem{miserendino2025swelancerfrontierllmsearn}
\bibinfo{author}{Samuel Miserendino}, \bibinfo{author}{Michele Wang}, \bibinfo{author}{Tejal Patwardhan}, et~al.
\newblock \bibinfo{title}{Swe-lancer: Can frontier llms earn \$1 million from real-world freelance software engineering?}, \bibinfo{year}{2025}.
\newblock \urlprefix\url{https://arxiv.org/abs/2502.12115}

\bibitem{tau-bench}
\bibinfo{author}{Shunyu Yao}, \bibinfo{author}{Noah Shinn}, \bibinfo{author}{Pedram Razavi}, et~al.
\newblock \bibinfo{title}{{\(\tau\)}-bench: {A} benchmark for tool-agent-user interaction in real-world domains}.
\newblock \bibinfo{journal}{CoRR}, \bibinfo{year}{2024}, \bibinfo{volume}{abs/2406.12045}.
\newblock \urlprefix\url{https://doi.org/10.48550/arXiv.2406.12045}

\bibitem{zhou2025sweetrltrainingmultiturnllm}
\bibinfo{author}{Yifei Zhou}, \bibinfo{author}{Song Jiang}, \bibinfo{author}{Yuandong Tian}, et~al.
\newblock \bibinfo{title}{Sweet-rl: Training multi-turn llm agents on collaborative reasoning tasks}, \bibinfo{year}{2025}.
\newblock \urlprefix\url{https://arxiv.org/abs/2503.15478}

\bibitem{wei2025browsecompsimplechallengingbenchmark}
\bibinfo{author}{Jason Wei}, \bibinfo{author}{Zhiqing Sun}, \bibinfo{author}{Spencer Papay}, et~al.
\newblock \bibinfo{title}{Browsecomp: A simple yet challenging benchmark for browsing agents}, \bibinfo{year}{2025}.
\newblock \urlprefix\url{https://arxiv.org/abs/2504.12516}

\bibitem{starace2025paperbenchevaluatingaisability}
\bibinfo{author}{Giulio Starace}, \bibinfo{author}{Oliver Jaffe}, \bibinfo{author}{Dane Sherburn}, et~al.
\newblock \bibinfo{title}{Paperbench: Evaluating ai's ability to replicate ai research}, \bibinfo{year}{2025}.
\newblock \urlprefix\url{https://arxiv.org/abs/2504.01848}

\bibitem{rein2025hcasthumancalibratedautonomysoftware}
\bibinfo{author}{David Rein}, \bibinfo{author}{Joel Becker}, \bibinfo{author}{Amy Deng}, et~al.
\newblock \bibinfo{title}{Hcast: Human-calibrated autonomy software tasks}, \bibinfo{year}{2025}.
\newblock \urlprefix\url{https://arxiv.org/abs/2503.17354}

\bibitem{wijk2024rebenchevaluatingfrontierai}
\bibinfo{author}{Hjalmar Wijk}, \bibinfo{author}{Tao Lin}, \bibinfo{author}{Joel Becker}, et~al.
\newblock \bibinfo{title}{Re-bench: Evaluating frontier ai r\&d capabilities of language model agents against human experts}, \bibinfo{year}{2024}.
\newblock \urlprefix\url{https://arxiv.org/abs/2411.15114}

\bibitem{Sutton2018}
\bibinfo{author}{Richard~S Sutton}, \bibinfo{author}{Andrew~G Barto}.
\newblock \bibinfo{title}{Reinforcement Learning: An Introduction}.
\newblock \bibinfo{edition}{Second} edition. \bibinfo{address}{Cambridge, MA, USA}: \bibinfo{publisher}{A Bradford Book, The MIT Press}. \bibinfo{year}{2018}, \bibinfo{year}{2018}.
\newblock \urlprefix\url{http://incompleteideas.net/book/the-book-2nd.html}

\bibitem{eysenbach2020supervised}
\bibinfo{author}{Ben Eysenbach}, \bibinfo{author}{Aviral Kumar}, \bibinfo{author}{Abhishek Gupta}.
\newblock \bibinfo{title}{Reinforcement learning is supervised learning on optimized data}, \bibinfo{year}{2020}.
\newblock \urlprefix\url{https://bair.berkeley.edu/blog/2020/10/13/supervised-rl/}, \bibinfo{note}{berkeley Artificial Intelligence Research Blog}

\bibitem{dsp}
\bibinfo{author}{Omar Khattab}, \bibinfo{author}{Keshav Santhanam}, \bibinfo{author}{Xiang~Lisa Li}, et~al.
\newblock \bibinfo{title}{Demonstrate-search-predict: Composing retrieval and language models for knowledge-intensive {NLP}}.
\newblock \bibinfo{journal}{CoRR}, \bibinfo{year}{2022}, \bibinfo{volume}{abs/2212.14024}.
\newblock \urlprefix\url{https://doi.org/10.48550/arXiv.2212.14024}

\bibitem{learnincontext}
\bibinfo{author}{Johannes Von~Oswald}, \bibinfo{author}{Eyvind Niklasson}, \bibinfo{author}{Ettore Randazzo}, et~al.
\newblock \bibinfo{title}{Transformers learn in-context by gradient descent}.
\newblock In: \bibinfo{booktitle}{Proceedings of the 40th International Conference on Machine Learning}, \bibinfo{publisher}{PMLR}. \bibinfo{year}{2023}, \emph{\bibinfo{series}{Proceedings of Machine Learning Research}}, volume \bibinfo{volume}{202}.
\newblock \bibinfo{pages}{35151--35174}.
\newblock \urlprefix\url{https://proceedings.mlr.press/v202/von-oswald23a.html}

\bibitem{reflexion}
\bibinfo{author}{Noah Shinn}, \bibinfo{author}{Federico Cassano}, \bibinfo{author}{Ashwin Gopinath}, et~al.
\newblock \bibinfo{title}{Reflexion: language agents with verbal reinforcement learning}.
\newblock In: \bibinfo{booktitle}{Advances in Neural Information Processing Systems 36: Annual Conference on Neural Information Processing Systems 2023, NeurIPS 2023, New Orleans, LA, USA, December 10 - 16, 2023}, \bibinfo{year}{2023}.
\newblock \urlprefix\url{http://papers.nips.cc/paper\_files/paper/2023/hash/1b44b878bb782e6954cd888628510e90-Abstract-Conference.html}

\bibitem{sumers2024cognitive}
\bibinfo{author}{Theodore Sumers}, \bibinfo{author}{Shunyu Yao}, \bibinfo{author}{Karthik Narasimhan}, et~al.
\newblock \bibinfo{title}{Cognitive architectures for language agents}.
\newblock \bibinfo{journal}{Transactions on Machine Learning Research}, \bibinfo{year}{2024}.
\newblock \urlprefix\url{https://openreview.net/forum?id=1i6ZCvflQJ}, \bibinfo{note}{survey Certification}

\bibitem{li2025surveyautomaticpromptengineering}
\bibinfo{author}{Wenwu Li}, \bibinfo{author}{Xiangfeng Wang}, \bibinfo{author}{Wenhao Li}, et~al.
\newblock \bibinfo{title}{A survey of automatic prompt engineering: An optimization perspective}, \bibinfo{year}{2025}.
\newblock \urlprefix\url{https://arxiv.org/abs/2502.11560}

\bibitem{promptbreeder}
\bibinfo{author}{Chrisantha Fernando}, \bibinfo{author}{Dylan Banarse}, \bibinfo{author}{Henryk Michalewski}, et~al.
\newblock \bibinfo{title}{Promptbreeder: Self-referential self-improvement via prompt evolution}.
\newblock In: \bibinfo{booktitle}{Forty-first International Conference on Machine Learning, {ICML} 2024, Vienna, Austria, July 21-27, 2024}, \bibinfo{publisher}{OpenReview.net}. \bibinfo{year}{2024}.
\newblock \urlprefix\url{https://openreview.net/forum?id=9ZxnPZGmPU}

\bibitem{evoprompt}
\bibinfo{author}{Qingyan Guo}, \bibinfo{author}{Rui Wang}, \bibinfo{author}{Junliang Guo}, et~al.
\newblock \bibinfo{title}{Connecting large language models with evolutionary algorithms yields powerful prompt optimizers}.
\newblock In: \bibinfo{booktitle}{The Twelfth International Conference on Learning Representations, {ICLR} 2024, Vienna, Austria, May 7-11, 2024}, \bibinfo{publisher}{OpenReview.net}. \bibinfo{year}{2024}.
\newblock \urlprefix\url{https://openreview.net/forum?id=ZG3RaNIsO8}

\bibitem{opro}
\bibinfo{author}{Chengrun Yang}, \bibinfo{author}{Xuezhi Wang}, \bibinfo{author}{Yifeng Lu}, et~al.
\newblock \bibinfo{title}{Large language models as optimizers}.
\newblock In: \bibinfo{booktitle}{The Twelfth International Conference on Learning Representations, {ICLR} 2024, Vienna, Austria, May 7-11, 2024}, \bibinfo{publisher}{OpenReview.net}. \bibinfo{year}{2024}.
\newblock \urlprefix\url{https://openreview.net/forum?id=Bb4VGOWELI}

\bibitem{sammo}
\bibinfo{author}{Tobias Schnabel}, \bibinfo{author}{Jennifer Neville}.
\newblock \bibinfo{title}{Prompts as programs: {A} structure-aware approach to efficient compile-time prompt optimization}.
\newblock \bibinfo{journal}{CoRR}, \bibinfo{year}{2024}, \bibinfo{volume}{abs/2404.02319}.
\newblock \urlprefix\url{https://doi.org/10.48550/arXiv.2404.02319}

\bibitem{abo}
\bibinfo{author}{Ruotian Ma}, \bibinfo{author}{Xiaolei Wang}, \bibinfo{author}{Xin Zhou}, et~al.
\newblock \bibinfo{title}{Are large language models good prompt optimizers?}
\newblock \bibinfo{journal}{CoRR}, \bibinfo{year}{2024}, \bibinfo{volume}{abs/2402.02101}.
\newblock \urlprefix\url{https://doi.org/10.48550/arXiv.2402.02101}

\bibitem{reprompt}
\bibinfo{author}{Weizhe Chen}, \bibinfo{author}{Sven Koenig}, \bibinfo{author}{Bistra Dilkina}.
\newblock \bibinfo{title}{Reprompt: Planning by automatic prompt engineering for large language models agents}, \bibinfo{year}{2024}.
\newblock \urlprefix\url{https://arxiv.org/abs/2406.11132}

\bibitem{gradsum}
\bibinfo{author}{Derek Austin}, \bibinfo{author}{Elliott Chartock}.
\newblock \bibinfo{title}{Grad-sum: Leveraging gradient summarization for optimal prompt engineering}, \bibinfo{year}{2024}.
\newblock \urlprefix\url{https://arxiv.org/abs/2407.12865}

\bibitem{self-discover}
\bibinfo{author}{Pei Zhou}, \bibinfo{author}{Jay Pujara}, \bibinfo{author}{Xiang Ren}, et~al.
\newblock \bibinfo{title}{Self-discover: Large language models self-compose reasoning structures}, \bibinfo{year}{2024}.
\newblock \urlprefix\url{https://arxiv.org/abs/2402.03620}

\bibitem{fu2024autoguide}
\bibinfo{author}{Yao Fu}, \bibinfo{author}{Dong-Ki Kim}, \bibinfo{author}{Jaekyeom Kim}, et~al.
\newblock \bibinfo{title}{Autoguide: Automated generation and selection of state-aware guidelines for large language model agents}.
\newblock \bibinfo{journal}{arXiv preprint arXiv:2403.08978}, \bibinfo{year}{2024}

\bibitem{automanual}
\bibinfo{author}{Minghao Chen}, \bibinfo{author}{Yihang Li}, \bibinfo{author}{Yanting Yang}, et~al.
\newblock \bibinfo{title}{Automanual: Generating instruction manuals by llm agents via interactive environmental learning}, \bibinfo{year}{2024}.
\newblock \urlprefix\url{https://arxiv.org/abs/2405.16247}

\bibitem{offlinetoonline}
\bibinfo{author}{Shenzhi Wang}, \bibinfo{author}{Qisen Yang}, \bibinfo{author}{Jiawei Gao}, et~al.
\newblock \bibinfo{title}{Train once, get a family: State-adaptive balances for offline-to-online reinforcement learning}.
\newblock In: \bibinfo{booktitle}{Advances in Neural Information Processing Systems}, \bibinfo{publisher}{Curran Associates, Inc.}. \bibinfo{year}{2023}, volume~\bibinfo{volume}{36}.
\newblock \bibinfo{pages}{47081--47104}.
\newblock \urlprefix\url{https://proceedings.neurips.cc/paper_files/paper/2023/file/9318763d049edf9a1f2779b2a59911d3-Paper-Conference.pdf}

\bibitem{Fu2024MSIAgentIM}
\bibinfo{author}{Dayuan Fu}, \bibinfo{author}{Biqing Qi}, \bibinfo{author}{Yihuai Gao}, et~al.
\newblock \bibinfo{title}{Msi-agent: Incorporating multi-scale insight into embodied agents for superior planning and decision-making}.
\newblock \bibinfo{journal}{arXiv preprint arXiv:2409.16686}, \bibinfo{year}{2024}.
\newblock \urlprefix\url{https://api.semanticscholar.org/CorpusID:272880637}

\bibitem{pouyanfar2018survey}
\bibinfo{author}{Samira Pouyanfar}, \bibinfo{author}{Saad Sadiq}, \bibinfo{author}{Yilin Yan}, et~al.
\newblock \bibinfo{title}{A survey on deep learning: Algorithms, techniques, and applications}.
\newblock \bibinfo{journal}{ACM Computing Surveys}, \bibinfo{year}{2018}, \bibinfo{volume}{51}: \bibinfo{pages}{1--36}

\bibitem{Shridhar2020ALFWorldAT}
\bibinfo{author}{Mohit Shridhar}, \bibinfo{author}{Xingdi Yuan}, \bibinfo{author}{Marc-Alexandre C{\^o}t{\'e}}, et~al.
\newblock \bibinfo{title}{Alfworld: Aligning text and embodied environments for interactive learning}.
\newblock \bibinfo{journal}{ArXiv}, \bibinfo{year}{2020}, \bibinfo{volume}{abs/2010.03768}.
\newblock \urlprefix\url{https://api.semanticscholar.org/CorpusID:222208810}

\bibitem{andreas-2022-language}
\bibinfo{author}{Jacob Andreas}.
\newblock \bibinfo{title}{Language models as agent models}.
\newblock In: \bibinfo{booktitle}{Findings of the Association for Computational Linguistics: EMNLP 2022}, \bibinfo{address}{Abu Dhabi, United Arab Emirates}: \bibinfo{publisher}{Association for Computational Linguistics}. \bibinfo{year}{2022}.
\newblock \bibinfo{pages}{5769--5779}.
\newblock \urlprefix\url{https://aclanthology.org/2022.findings-emnlp.423/}

\bibitem{agentoptimizer}
\bibinfo{author}{Shaokun Zhang}, \bibinfo{author}{Jieyu Zhang}, \bibinfo{author}{Jiale Liu}, et~al.
\newblock \bibinfo{title}{Offline training of language model agents with functions as learnable weights}.
\newblock In: \bibinfo{booktitle}{Forty-first International Conference on Machine Learning, {ICML} 2024, Vienna, Austria, July 21-27, 2024}, \bibinfo{publisher}{OpenReview.net}. \bibinfo{year}{2024}.
\newblock \urlprefix\url{https://openreview.net/forum?id=2xbkWiEuR1}

\bibitem{learnact}
\bibinfo{author}{Haiteng Zhao}, \bibinfo{author}{Chang Ma}, \bibinfo{author}{Guoyin Wang}, et~al.
\newblock \bibinfo{title}{Empowering large language model agents through action learning}.
\newblock \bibinfo{journal}{CoRR}, \bibinfo{year}{2024}, \bibinfo{volume}{abs/2402.15809}.
\newblock \urlprefix\url{https://doi.org/10.48550/arXiv.2402.15809}

\bibitem{wang2024voyager}
\bibinfo{author}{Guanzhi Wang}, \bibinfo{author}{Yuqi Xie}, \bibinfo{author}{Yunfan Jiang}, et~al.
\newblock \bibinfo{title}{Voyager: An open-ended embodied agent with large language models}.
\newblock \bibinfo{journal}{Transactions on Machine Learning Research}, \bibinfo{year}{2024}.
\newblock \urlprefix\url{https://openreview.net/forum?id=ehfRiF0R3a}

\bibitem{wang2024agentworkflowmemory}
\bibinfo{author}{Zora~Zhiruo Wang}, \bibinfo{author}{Jiayuan Mao}, \bibinfo{author}{Daniel Fried}, et~al.
\newblock \bibinfo{title}{Agent workflow memory}, \bibinfo{year}{2024}.
\newblock \urlprefix\url{https://arxiv.org/abs/2409.07429}

\bibitem{moss}
\bibinfo{author}{Andrew Zhao}, \bibinfo{author}{Matthieu Lin}, \bibinfo{author}{Yangguang Li}, et~al.
\newblock \bibinfo{title}{A mixture of surprises for unsupervised reinforcement learning}.
\newblock In: \bibinfo{booktitle}{Advances in Neural Information Processing Systems}, \bibinfo{publisher}{Curran Associates, Inc.}. \bibinfo{year}{2022}, volume~\bibinfo{volume}{35}.
\newblock \bibinfo{pages}{26078--26090}.
\newblock \urlprefix\url{https://proceedings.neurips.cc/paper_files/paper/2022/file/a7667ee5d545a43d2f0fda98863c260e-Paper-Conference.pdf}

\bibitem{eysenbach2018diversity}
\bibinfo{author}{Benjamin Eysenbach}, \bibinfo{author}{Abhishek Gupta}, \bibinfo{author}{Julian Ibarz}, et~al.
\newblock \bibinfo{title}{Diversity is all you need: Learning skills without a reward function}.
\newblock In: \bibinfo{booktitle}{International Conference on Learning Representations}, \bibinfo{year}{2019}.
\newblock \urlprefix\url{https://openreview.net/forum?id=SJx63jRqFm}

\bibitem{liu2025contextual}
\bibinfo{author}{Yitao Liu}, \bibinfo{author}{Chenglei Si}, \bibinfo{author}{Karthik~R Narasimhan}, et~al.
\newblock \bibinfo{title}{Contextual experience replay for continual learning of language agents}, \bibinfo{year}{2025}.
\newblock \urlprefix\url{https://openreview.net/forum?id=RXvFK5dnpz}

\bibitem{ovadia2019trustmodelsuncertaintyevaluating}
\bibinfo{author}{Yaniv Ovadia}, \bibinfo{author}{Emily Fertig}, \bibinfo{author}{Jie Ren}, et~al.
\newblock \bibinfo{title}{Can you trust your model's uncertainty? evaluating predictive uncertainty under dataset shift}, \bibinfo{year}{2019}.
\newblock \urlprefix\url{https://arxiv.org/abs/1906.02530}

\bibitem{zheng2025skillweaverwebagentsselfimprove}
\bibinfo{author}{Boyuan Zheng}, \bibinfo{author}{Michael~Y Fatemi}, \bibinfo{author}{Xiaolong Jin}, et~al.
\newblock \bibinfo{title}{Skillweaver: Web agents can self-improve by discovering and honing skills}, \bibinfo{year}{2025}.
\newblock \urlprefix\url{https://arxiv.org/abs/2504.07079}

\bibitem{paszke2019pytorch}
\bibinfo{author}{Adam Paszke}, \bibinfo{author}{Sam Gross}, \bibinfo{author}{Francisco Massa}, et~al.
\newblock \bibinfo{title}{Pytorch: An imperative style, high-performance deep learning library}.
\newblock \bibinfo{journal}{Advances in Neural Information Processing Systems}, \bibinfo{year}{2019}, \bibinfo{volume}{32}.
\newblock \urlprefix\url{https://papers.nips.cc/paper/2019/hash/bdbca288fee7f92f2bfa9f7012727740-Abstract.html}

\bibitem{symboliclearning}
\bibinfo{author}{Wangchunshu Zhou}, \bibinfo{author}{Yixin Ou}, \bibinfo{author}{Shengwei Ding}, et~al.
\newblock \bibinfo{title}{Symbolic learning enables self-evolving agents}.
\newblock \bibinfo{journal}{CoRR}, \bibinfo{year}{2024}, \bibinfo{volume}{abs/2406.18532}.
\newblock \urlprefix\url{https://doi.org/10.48550/arXiv.2406.18532}

\bibitem{wang2025how}
\bibinfo{author}{Wenyi Wang}, \bibinfo{author}{Hisham~Abdullah Alyahya}, \bibinfo{author}{Dylan~R Ashley}, et~al.
\newblock \bibinfo{title}{How to correctly do semantic backpropagation on language-based agentic systems}, \bibinfo{year}{2025}.
\newblock \urlprefix\url{https://openreview.net/forum?id=r1cbFEH0Df}

\bibitem{van-duijn-etal-2023-theory}
\bibinfo{author}{Max van Duijn}, \bibinfo{author}{Bram van Dijk}, \bibinfo{author}{Tom Kouwenhoven}, et~al.
\newblock \bibinfo{title}{Theory of mind in large language models: Examining performance of 11 state-of-the-art models vs. children aged 7–10 on advanced tests}.
\newblock In: \bibinfo{booktitle}{Proceedings of the 27th Conference on Computational Natural Language Learning (CoNLL)}, \bibinfo{address}{Singapore}: \bibinfo{publisher}{Association for Computational Linguistics}. \bibinfo{year}{2023}.
\newblock \bibinfo{pages}{389--402}.
\newblock \urlprefix\url{https://aclanthology.org/2023.conll-1.25/}

\bibitem{creditassignment}
\bibinfo{author}{Eduardo Pignatelli}, \bibinfo{author}{Johan Ferret}, \bibinfo{author}{Matthieu Geist}, et~al.
\newblock \bibinfo{title}{A survey of temporal credit assignment in deep reinforcement learning}.
\newblock \bibinfo{journal}{Transactions on Machine Learning Research}, \bibinfo{year}{2024}.
\newblock \urlprefix\url{https://openreview.net/forum?id=bNtr6SLgZf}, \bibinfo{note}{survey Certification}

\bibitem{kwa2025measuringaiabilitycomplete}
\bibinfo{author}{Thomas Kwa}, \bibinfo{author}{Ben West}, \bibinfo{author}{Joel Becker}, et~al.
\newblock \bibinfo{title}{Measuring ai ability to complete long tasks}, \bibinfo{year}{2025}.
\newblock \urlprefix\url{https://arxiv.org/abs/2503.14499}

\bibitem{retroformer}
\bibinfo{author}{Weiran Yao}, \bibinfo{author}{Shelby Heinecke}, \bibinfo{author}{Juan~Carlos Niebles}, et~al.
\newblock \bibinfo{title}{Retroformer: Retrospective large language agents with policy gradient optimization}.
\newblock In: \bibinfo{booktitle}{The Twelfth International Conference on Learning Representations, {ICLR} 2024, Vienna, Austria, May 7-11, 2024}, \bibinfo{publisher}{OpenReview.net}. \bibinfo{year}{2024}.
\newblock \urlprefix\url{https://openreview.net/forum?id=KOZu91CzbK}

\bibitem{agentsurvey}
\bibinfo{author}{Bang Liu}, \bibinfo{author}{Xinfeng Li}, \bibinfo{author}{Jiayi Zhang}, et~al.
\newblock \bibinfo{title}{Advances and challenges in foundation agents: From brain-inspired intelligence to evolutionary, collaborative, and safe systems}, \bibinfo{year}{2025}.
\newblock \urlprefix\url{https://arxiv.org/abs/2504.01990}

\bibitem{openai2025memory}
\bibinfo{author}{OpenAI}.
\newblock \bibinfo{title}{Openai memory announcement}, \bibinfo{year}{2025}.
\newblock \urlprefix\url{https://x.com/OpenAI/status/1910378768172212636}, \bibinfo{note}{accessed: 2025-05-13}

\bibitem{suzgun2025dynamiccheatsheettesttimelearning}
\bibinfo{author}{Mirac Suzgun}, \bibinfo{author}{Mert Yuksekgonul}, \bibinfo{author}{Federico Bianchi}, et~al.
\newblock \bibinfo{title}{Dynamic cheatsheet: Test-time learning with adaptive memory}, \bibinfo{year}{2025}.
\newblock \urlprefix\url{https://arxiv.org/abs/2504.07952}

\bibitem{sutton2019bitter}
\bibinfo{author}{Richard~S Sutton}.
\newblock \bibinfo{title}{The bitter lesson}, \bibinfo{year}{2019}.
\newblock \urlprefix\url{http://www.incompleteideas.net/IncIdeas/BitterLesson.html}, \bibinfo{note}{accessed: 2025-05-12}

\bibitem{streambench}
\bibinfo{author}{Cheng-Kuang Wu}, \bibinfo{author}{Zhi~Rui Tam}, \bibinfo{author}{Chieh-Yen Lin}, et~al.
\newblock \bibinfo{title}{Streambench: Towards benchmarking continuous improvement of language agents}.
\newblock In: \bibinfo{booktitle}{Advances in Neural Information Processing Systems}, \bibinfo{publisher}{Curran Associates, Inc.}. \bibinfo{year}{2024}, volume~\bibinfo{volume}{37}.
\newblock \bibinfo{pages}{107039--107063}.
\newblock \urlprefix\url{https://proceedings.neurips.cc/paper_files/paper/2024/file/c189915371c4474fe9789be3728113fc-Paper-Datasets_and_Benchmarks_Track.pdf}

\bibitem{llmevolve}
\bibinfo{author}{Jiaxuan You}, \bibinfo{author}{Mingjie Liu}, \bibinfo{author}{Shrimai Prabhumoye}, et~al.
\newblock \bibinfo{title}{Llm-evolve: Evaluation for llm's evolving capability on benchmarks}.
\newblock In: \bibinfo{booktitle}{Proceedings of the 2024 Conference on Empirical Methods in Natural Language Processing}, \bibinfo{year}{2024}.
\newblock \bibinfo{pages}{16937--16942}.
\newblock \urlprefix\url{https://aclanthology.org/2024.emnlp-main.940}

\bibitem{lin2025sleeptimecomputeinferencescaling}
\bibinfo{author}{Kevin Lin}, \bibinfo{author}{Charlie Snell}, \bibinfo{author}{Yu~Wang}, et~al.
\newblock \bibinfo{title}{Sleep-time compute: Beyond inference scaling at test-time}, \bibinfo{year}{2025}.
\newblock \urlprefix\url{https://arxiv.org/abs/2504.13171}

\bibitem{nikishin2022primacybiasdeepreinforcement}
\bibinfo{author}{Evgenii Nikishin}, \bibinfo{author}{Max Schwarzer}, \bibinfo{author}{Pierluca D'Oro}, et~al.
\newblock \bibinfo{title}{The primacy bias in deep reinforcement learning}, \bibinfo{year}{2022}.
\newblock \urlprefix\url{https://arxiv.org/abs/2205.07802}

\bibitem{achille2019criticallearningperiodsdeep}
\bibinfo{author}{Alessandro Achille}, \bibinfo{author}{Matteo Rovere}, \bibinfo{author}{Stefano Soatto}.
\newblock \bibinfo{title}{Critical learning periods in deep neural networks}, \bibinfo{year}{2019}.
\newblock \urlprefix\url{https://arxiv.org/abs/1711.08856}

\bibitem{frankle2020earlyphaseneuralnetwork}
\bibinfo{author}{Jonathan Frankle}, \bibinfo{author}{David~J Schwab}, \bibinfo{author}{Ari~S Morcos}.
\newblock \bibinfo{title}{The early phase of neural network training}, \bibinfo{year}{2020}.
\newblock \urlprefix\url{https://arxiv.org/abs/2002.10365}

\bibitem{elsayed2024streamingdeepreinforcementlearning}
\bibinfo{author}{Mohamed Elsayed}, \bibinfo{author}{Gautham Vasan}, \bibinfo{author}{A~Rupam Mahmood}.
\newblock \bibinfo{title}{Streaming deep reinforcement learning finally works}, \bibinfo{year}{2024}.
\newblock \urlprefix\url{https://arxiv.org/abs/2410.14606}

\bibitem{karami2025latticelearningefficientlycompress}
\bibinfo{author}{Mahdi Karami}, \bibinfo{author}{Vahab Mirrokni}.
\newblock \bibinfo{title}{Lattice: Learning to efficiently compress the memory}, \bibinfo{year}{2025}.
\newblock \urlprefix\url{https://arxiv.org/abs/2504.05646}

\bibitem{lee2024aligning}
\bibinfo{author}{Seongyun Lee}, \bibinfo{author}{Sue~Hyun Park}, \bibinfo{author}{Seungone Kim}, et~al.
\newblock \bibinfo{title}{Aligning to thousands of preferences via system message generalization}.
\newblock In: \bibinfo{booktitle}{The Thirty-eighth Annual Conference on Neural Information Processing Systems}, \bibinfo{year}{2024}.
\newblock \urlprefix\url{https://openreview.net/forum?id=recsheQ7e8}

\bibitem{tan2025langprobelanguageprogramsbenchmark}
\bibinfo{author}{Shangyin Tan}, \bibinfo{author}{Lakshya~A Agrawal}, \bibinfo{author}{Arnav Singhvi}, et~al.
\newblock \bibinfo{title}{Langprobe: a language programs benchmark}, \bibinfo{year}{2025}.
\newblock \urlprefix\url{https://arxiv.org/abs/2502.20315}

\bibitem{jiang2023vima}
\bibinfo{author}{Yunfan Jiang}, \bibinfo{author}{Agrim Gupta}, \bibinfo{author}{Zichen Zhang}, et~al.
\newblock \bibinfo{title}{Vima: General robot manipulation with multimodal prompts}.
\newblock In: \bibinfo{booktitle}{Fortieth International Conference on Machine Learning}, \bibinfo{year}{2023}

\end{thebibliography}

\end{document}